\documentclass[10pt,twocolumn,letterpaper]{article}

\usepackage{wacv}              

\usepackage{booktabs}
\usepackage{multirow}

\usepackage{amssymb}
\usepackage{amsmath}

\usepackage{booktabs}
\usepackage{graphicx}
\usepackage{amsmath}
\usepackage{amssymb}
\usepackage[table]{xcolor}
\usepackage{multirow}
\usepackage{booktabs}
\usepackage{graphicx}
\usepackage{adjustbox}
\usepackage{bm} 
\usepackage{fix-cm} 

\usepackage{algorithm}
\usepackage{algorithmic}

\usepackage[table]{xcolor}
\definecolor{newrow}{gray}{0.93}
\definecolor{wacvblue}{rgb}{0.21,0.49,0.74}
\usepackage[pagebackref,breaklinks,colorlinks,allcolors=wacvblue]{hyperref}

\def\wacvPaperID{1478} 
\def\confName{WACV}
\def\confYear{2027}

\title{Attention Drift Regularization with Adaptive Initialization for Domain-Robust Visual Recognition}

\author{Shivangi Rai, Vinod K Kurmi\\
Indian Institute of Science Education and Research Bhopal, India\\
{\tt\small \{shivangir23, vinodkk\}@iiserb.ac.in}\\}

\begin{document}
\maketitle
%
%
%


\begin{abstract}
Source-free domain adaptation~(SFDA) adapts a pre-trained source
model to unlabeled target domains without source data access, yet two
critical failure modes remain unaddressed.
First, \textit{attention drift}: source-learned attention maps shift
toward spurious or background features under distribution shift,
destabilizing feature representations.
Second, \textit{domain-gap-agnostic initialization}: applying a single
fixed source model uniformly across all target domains ignores domain
gap severity, causing premature overfitting to noisy pseudo-labels
under large shifts.
We propose \textbf{DGA-CAFC} to address both failures.
The \textit{Counterfactual Attention-based Feature Consistency}
(\textbf{CAFC}) module enforces feature stability by minimizing the
discrepancy between representations computed under learned attention
and those under counterfactual-inspired normalized perturbations, thereby
promoting domain-robust features independent of spurious attention shifts.
The \textit{Domain-Gap Aware Source Model Selection}
(\textbf{DGASS}) module estimates domain gap severity via target
prediction entropy and selects the source model variant whose
confidence level best matches the estimated gap, reducing negative
transfer and stabilizing pseudo-label quality.
Crucially, CAFC transfers beyond SFDA: applied as a lightweight
regularizer to standard domain generalization objectives without
target data. Extensive experiments across SFDA and DG benchmarks demonstrate that our method consistently improves over recent baselines, demonstrating
the effectiveness and generality of the proposed framework. Project page is available at \url{https://visdomlab.github.io/DGA_CAFC}

%
\end{abstract}

\section{Introduction}
\label{sec:intro}

Visual recognition models trained on a labelled source domain
degrade when deployed under distribution shift caused by
differences in imaging conditions, object appearance, texture,
background composition, or acquisition device.
Two settings address this without full re-annotation of the
target domain.
In \textit{source-free domain adaptation} (SFDA)~%
\cite{liang2020we,yang2021exploiting,yang2022attracting,lee2022confidence},
a source-trained model and unlabeled target data are available
at adaptation time, but the source training set is inaccessible
due to privacy or storage constraints.
In \textit{domain generalization}
(DG)~\cite{gulrajani2020search,sagawa2019distributionally},
no target data are available during training at all; the model
must generalize to unseen domains at test time.
Both settings demand representations that are stable under
covariate shift, yet the mechanisms by which standard
training produces fragile representations remain insufficiently
characterized.

Transformer-based backbones~\cite{dosovitskiy2020image,liu2021swin}
dominate visual recognition, including domain adaptive
models~\cite{yang2021transformer,sun2022safe}.
Their self-attention mechanism aggregates features from all
spatial tokens, weighting each by a learned attention map
derived from source-domain statistics.
Under target-domain shift, these attention maps can drift
toward spurious or background features that are discriminative
in the source domain but not in the target,
a phenomenon we term \textit{attention drift}.
Attention drift destabilizes the attention-conditioned
feature representations on which downstream tasks
depends: in SFDA, where a teacher model generates pseudo-labels
from these representations, corrupted features propagate
errors into the pseudo-label signal and compound through
self-training; in DG, drifted attention patterns reduce
robustness to the appearance variations encountered at test
time on unseen domains.
Despite the broad adoption of attention-based backbones, existing
SFDA and DG methods do not directly target the stability
of attention-conditioned features under distribution shift.
A second, SFDA-specific failure mode arises at initialisation.
Standard SFDA methods~\cite{liang2020we,yang2021transformer}
initialise the student model from a single fixed source
checkpoint regardless of the severity of the domain gap
between source and target.
This is suboptimal because gap severity varies substantially
across transfer tasks: a high-gap pair requires a conservative
initialisation that does not over-commit to source-domain
decision boundaries, whereas a low-gap pair benefits from
a confident initialisation that accelerates pseudo-label
convergence.
Ignoring this distinction causes premature overfitting to
noisy pseudo-labels under large domain shifts, degrading
adaptation quality from the first training step.

\begin{figure}[t]
    \begin{center}
    \includegraphics[width=1\linewidth]{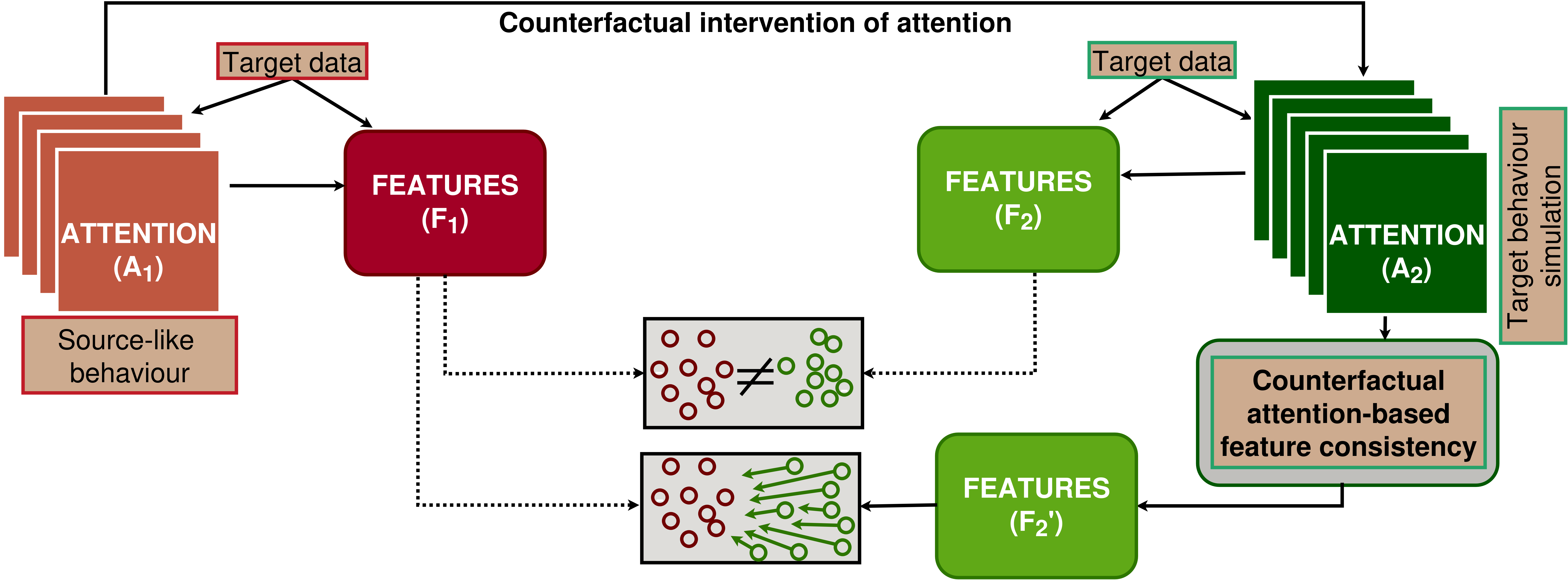}
   \caption{Counterfactual Attention-based Feature Consistency (CAFC).
CAFC enforces feature-level consistency between representations
computed under learned attention and counterfactual attention
perturbations, encouraging robustness to attention drift.}
    \label{fig:cafc}
    \end{center}
    \vspace{-2.5em}
\end{figure}

\begin{figure}[t]
    \centering
    \includegraphics[width=\columnwidth]{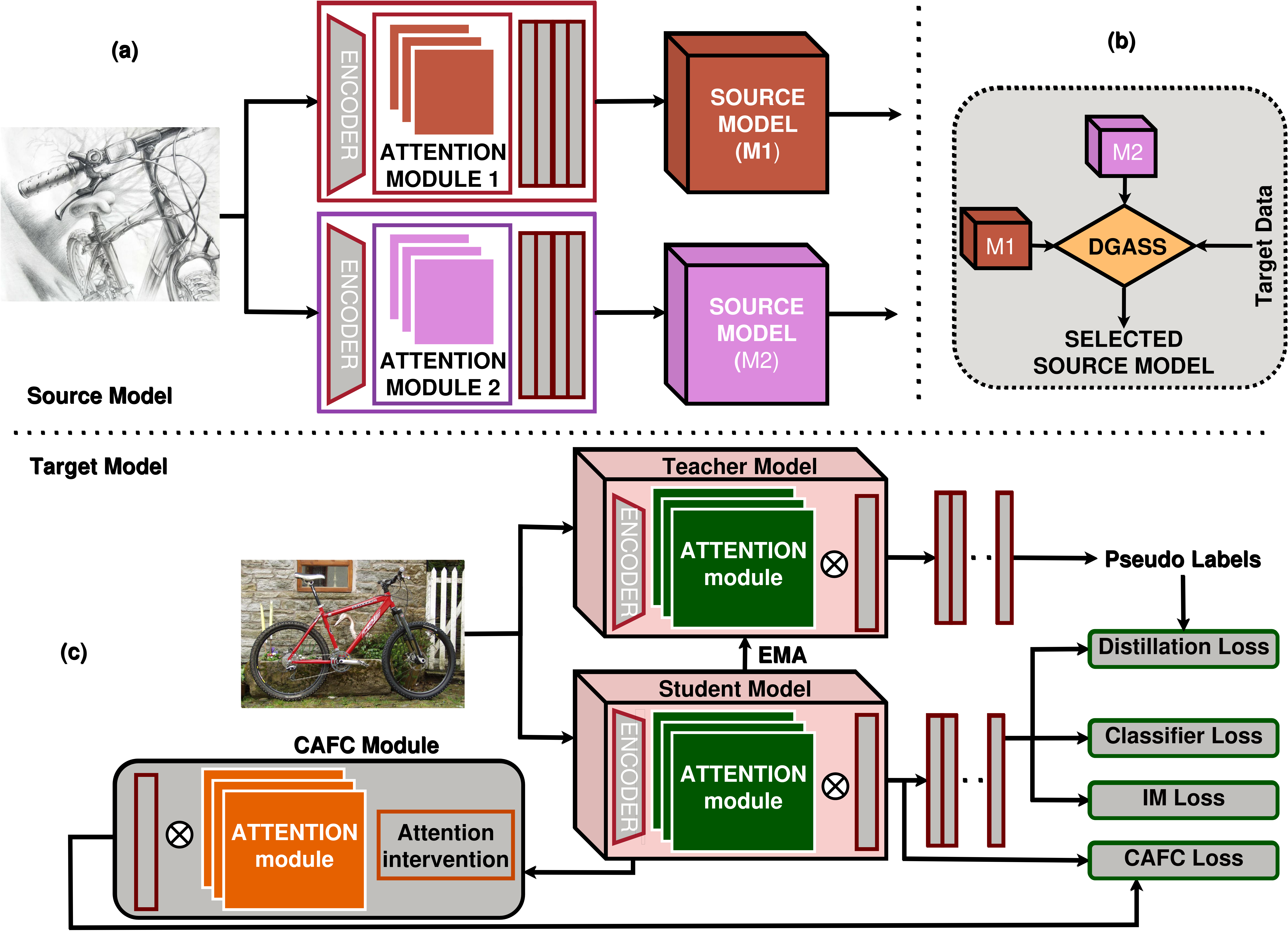}
    \caption{Overview of the proposed DGA-CAFC framework.
Source training produces two source candidates with complementary
attention behavior. DGASS selects the source candidate according
to target prediction entropy, and CAFC regularizes target adaptation
by enforcing consistency between learned and counterfactual
attention-conditioned features (best view in color).}
    \label{fig:complete_framework}
    \vspace{-1.5em} 
\end{figure}

We propose \textbf{DGA-CAFC}, a two-module framework that
addresses both failure modes (\textbf{Fig.~\ref{fig:complete_framework}}).
The \textit{Counterfactual Attention-based Feature Consistency}
(\textbf{CAFC}) module enforces stability in
attention-conditioned features by minimizing the discrepancy
between representations computed under the model's learned
attention map and those computed under three
counterfactual-inspired normalised attention perturbations:
a uniform map that removes all focal structure, a random
map that injects noise, and a reversed map that inverts
saliency.
Minimising this discrepancy is analogous to Lipschitz-style
regularization in attention-conditioned feature space: it
reduces the sensitivity of the learned representation to the
specific attention weight configuration, encouraging
domain-robust features that do not depend on
source-specific attention patterns.
Critically, CAFC requires no target labels and operates
on attention outputs, making it directly applicable in both
SFDA and DG settings without architectural modification.
The \textit{Domain-Gap Aware Source Model Selection}
(\textbf{DGASS}) module addresses the SFDA initialisation
problem by estimating target domain gap severity from
unlabeled target prediction entropy and selecting the
source model variant between a sharp, confident variant
and a smoother, more conservative one.
Our contributions can be summarized as follows:

\begin{enumerate}

 \item We propose \textbf{CAFC}, an attention-stability
  regularizer that operates without target labels and
  transfers across settings, functioning within the SFDA
  adaptation and as a source-training regularizer
  in domain generalization.

  \item We propose \textbf{DGASS}, an entropy-guided
  source model selection mechanism for SFDA that requires
  no target labels and acts before adaptation begins,
  selecting the source variant whose confidence
  calibration matches the estimated domain gap.

  \item We extend CAFC to domain generalization as a
  lightweight source-training regularizer, demonstrating
  that attention-stability regularization improves
  multiple standard DG objectives
  without requiring any target data during training.

  \item Extensive experiments and analysis are conducted
across multiple SFDA and  DG benchmarks,
with thorough ablation studies provided to justify
each design choice and validate the contribution of
each proposed module.


\end{enumerate}


\section{Related Work}
\label{sec:related}

\noindent\textbf{Source-Free Domain Adaptation.}
SFDA adapts a source-trained model to an unlabelled target
without source data~\cite{liang2020we}.
Pseudo-label refinement methods such as
SHOT~\cite{liang2020we}, NRC~\cite{yang2021exploiting},
AaD~\cite{yang2022attracting}, and
CoWA-JMDS~\cite{lee2022confidence} improve adaptation through
information maximisation, neighbourhood clustering, or
confidence-weighted alignment.
Transformer-based methods including
TransDA~\cite{yang2021transformer},
TVT~\cite{yang2108tvt}, SSRT~\cite{sun2022safe}, leverage attention mechanisms for
stronger feature transfer.
None of these methods directly target \textit{attention drift}, i.e.,
the instability of attention-conditioned features under
distribution shift nor do they account for domain gap
severity at source model initialization.

\noindent\textbf{Attention in Domain Adaptation.}
Transformer backbones~\cite{dosovitskiy2020image,liu2021swin}
now underpin many domain-adaptive
models~\cite{yang2021transformer,sun2022safe},
with cross-attention and patch-mix strategies exploited for
cross-domain
transfer~\cite{xu2021cdtrans,zhu2023patch}.
These approaches utilise attention for domain transfer but
do not regularise the stability of attention-conditioned
representations when source-learned attention maps are
applied to out-of-distribution target data.
\noindent\textbf{Counterfactual and Causal Learning.}
Counterfactual attention has been applied to fine-grained
recognition by evaluating causal relationships between
attention regions and
predictions~\cite{rao2021counterfactual,yu2025attention}.
Unlike these methods, which refine attention for
within-domain recognition, CAFC uses counterfactual-inspired
perturbations as stress tests for cross-domain feature
stability, operating without any target supervision.
\noindent\textbf{Domain Generalization.}
DG methods train solely on source domains and evaluate on
unseen targets~\cite{gulrajani2020search}.
Representative objectives include feature alignment
(CORAL~\cite{sun2016deep}, MMD~\cite{volpi2018generalizing}),
interpolation regularisation
(Mixup~\cite{zhang2017mixup}), and distributionally robust
optimisation~\cite{sagawa2019distributionally,krueger2021out}.
None impose constraints on attention stability;
CAFC is orthogonal to each and can be added as a
lightweight source-training regularizer without
modifying the base objective.

\section{Methodology}
\label{sec:method}

\subsection{Problem Formulation}
\label{sec:method_problem}

Let $f_s$ denote a source model pre-trained on a labeled source
domain $\mathcal{D}_s$ and frozen thereafter.
In source-free domain adaptation (SFDA), a student model $f_t$
is initialised from $f_s$ and adapted to an unlabeled target
domain $\mathcal{D}_t = \{x_t\}$ without accessing $\mathcal{D}_s$.
An EMA teacher $f_{\mathrm{ema}}$ provides soft supervision.
The parameters  $\theta_{\mathrm{ema}}$ are updated as a
moving average of ${\theta}_t$, yielding temporally stable
pseudo-label predictions during adaptation.
The adaptation objective is:
\begin{equation}
  \label{eq:objective}
  f_t^* \;=\;
  \operatorname*{arg\,min}_{f_t}
  \;\mathcal{L}_{\mathrm{total}}
  \!\left(f_t;\,\mathcal{D}_t,\,f_{\mathrm{ema}}\right)
\end{equation}
where $\mathcal{L}_{\mathrm{total}}$ is fully defined in
Sec.~\ref{sec:method_training}.
We denote the $C$-class softmax output at target sample $x_t$
as $\mathbf{p}(x_t) \in \Delta^{C-1}$.

\subsection{Framework Overview}
\label{sec:method_overview}

\textbf{DGA-CAFC} addresses two distinct failure modes of SFDA
through two complementary modules (\textbf{Fig.~\ref{fig:complete_framework}}).
\textit{Attention drift}, wherein source-learned attention maps
shift toward spurious target-domain features is addressed by
\textbf{CAFC} (\textbf{Fig. \ref{fig:cafc}) (Sec.~\ref{sec:method_cafc}}), which regularises
attention-conditioned features throughout adaptation.
\textit{Domain-gap-agnostic initialisation}, wherein a single
fixed source model is applied uniformly regardless of domain gap
severity, it is addressed by \textbf{DGASS}
(Sec.~\ref{sec:method_dgass}), which selects the source variant
best matched to the target domain's estimated gap at
initialization time.
The two modules are architecturally independent: DGASS acts
once before training begins whereas CAFC acts at every gradient step.

\subsection{Counterfactual Attention-based Feature Consistency}
\label{sec:method_cafc}

\noindent\textbf{Learned attention and features.}
Let $\mathbf{Q}, \mathbf{K}, \mathbf{V}
\in \mathbb{R}^{N \times d}$ be the query, key, and value
matrices of a self-attention layer over $N$ tokens with
dimension $d$.
The learned attention map and its conditioned feature are:
%
%
\begin{equation}
\vspace{-0.7em}
\label{eq:attn_map}
\resizebox{0.4\columnwidth}{!}{$\displaystyle
\mathbf{A}_{L}
=
\mathrm{softmax}\!\left(
\frac{\mathbf{Q}\mathbf{K}^{\top}}{\sqrt{d}}
\right)
$}
\end{equation}
\begin{equation}
\vspace{-0.5em}
\label{eq:attn_feature}
\mathbf{h}_{L}=\mathbf{A}_{L}\mathbf{V}
\end{equation}
where each row of $\mathbf{A}_{L}$ is a valid probability
distribution over $N$ tokens.
Under distribution shift, $\mathbf{A}_{L}$ drifts toward
target-domain-specific spurious cues, making $\mathbf{h}_{L}$
fragile.

\noindent\textbf{Counterfactual-inspired perturbations.}
We define three structured attention perturbations, each
stressing a distinct type of attention dependency:
%


\begin{equation}
\label{eq:perturb}
\makebox[\columnwidth][l]{%
\resizebox{0.98\columnwidth}{!}{$\displaystyle
\begin{array}{@{}lcll@{}}
\mathbf{A}_{\mathrm{unif}}[i,j]
&=&
\dfrac{1}{N},\quad \forall\,i,j,
& \text{removes focal structure},\\[4pt]

\mathbf{A}_{\mathrm{rand}}
&=&
\mathrm{Norm}(\mathbf{E}),\quad
\mathbf{E}\sim\mathcal{U}(0,1),
& \text{injects noisy attention},\\[4pt]

\mathbf{A}_{\mathrm{rev}}
&=&
\mathrm{Norm}\!\left(
\max\!\left(1-\mathbf{A}_{L},\mathbf{0}\right)
\right),
& \text{reverses salient regions}.
\end{array}
$}%
}
\end{equation}
%



where $\mathrm{Norm}(\cdot)$ denotes row-wise
$\ell_1$-normalisation, ensuring each map is a valid
probability distribution (necessity established in
Supplementary). The three perturbations serve as complementary stress tests:
$\mathbf{A}_{\mathrm{unif}}$ removes all discriminative focus,
testing whether features remain stable without any attentional
guidance; $\mathbf{A}_{\mathrm{rand}}$ injects noisy attention,
testing stability under unreliable focus; and
$\mathbf{A}_{\mathrm{rev}}$ inverts saliency, testing robustness
against over-reliance on salient but potentially spurious
source-domain regions.
Averaging these three maps into $\mathbf{A}_{cf}$ produces a
maximally uninformative attention distribution with respect
to the learned $\mathbf{A}_L$, ensuring that
$\mathcal{L}_{\mathrm{CAFC}}$ targets the full complement of
attention-drift failure modes.
The averaged counterfactual attention map and its feature are:
%
\begin{equation}
\label{eq:acf}
\scalebox{0.85}{$\displaystyle
\mathbf{A}_{cf}
=
\mathrm{}\!\left(
\frac{
\mathbf{A}_{\mathrm{unif}}
+
\mathbf{A}_{\mathrm{rand}}
+
\mathbf{A}_{\mathrm{rev}}
}{3}
\right)
$}
\end{equation}
\begin{equation}
\label{eq:hcf}
\mathbf{h}_{cf}
=
\mathbf{A}_{cf}\mathbf{V}
\end{equation}
%
Minimising the discrepancy between $\mathbf{h}_{L}$ and
$\mathbf{h}_{\mathrm{cf}}$ reduces the sensitivity of the
representation to the specific attention configuration. This can
be viewed as a form of attention-conditional Lipschitz
regularization, promoting features that remain stable under
attention drift across domains.
The \textbf{CAFC loss} is defined as:
\begin{equation}
  \label{eq:lcafc}
  \mathcal{L}_{\mathrm{CAFC}}
  = \left\|\,\mathbf{h}_{L} - \mathbf{h}_{cf}\,\right\|_{2}^{2}
\end{equation}
$\mathcal{L}_{\mathrm{CAFC}}$ operates on the target domain
without labels, making it applicable in both SFDA and DG settings. The design rationale for these perturbations as complementary attention-space stress tests is provided in the Supplementary.


\subsection{Domain-Gap Aware Source Model Selection}
\label{sec:method_dgass}
We propose \textbf{DGASS}, a lightweight entropy-guided strategy that selects the source model variant best matched to the estimated severity of the target domain gap, providing a stronger initialisation starting point for SFDA. \\
\noindent\textbf{Source model variants.}
We construct source candidates by varying only the
self-attention softmax formulation. Let
$e_{ij}=\mathbf{q}_i\cdot\mathbf{k}_j/\sqrt{d}$
denote the raw attention score between tokens $i$ and $j$.
$f_s^1$ and $f_s^2$ denote the two source candidates used by DGASS.
The first source candidate, $f_s^1$, uses standard
scaled-dot-product softmax attention (eq. \ref{eq:fs1}):
\begin{equation}
  \label{eq:fs1}
  \alpha^{(1)}_{ij}
  =
  \frac{\exp(e_{ij})}
       {\sum_{j'}\exp(e_{ij'})}
\end{equation}
This formulation produces sharp and confident attention
distributions. We also define an auxiliary weighted-softmax formulation
$\alpha^{(2)}_{ij}$ using min--max rescaling. First, the
rescaling weight is computed as (eq. \ref{eq:omega})
\begin{equation}
  \label{eq:omega}
  \omega_{ij}
  =
  \frac{
    e_{ij}-\min_{j'} e_{ij'}
  }{
    \max_{j'} e_{ij'}-\min_{j'} e_{ij'}+\epsilon
  } 
\end{equation}
\begin{equation}
  \label{eq:fs2}
  \alpha^{(2)}_{ij}
  =
  \omega_{ij}
  \frac{\exp(e_{ij})}
       {\sum_{j'}\exp(e_{ij'})}
\end{equation}
Here, $\epsilon>0$ is a small constant for numerical
stability. This weighted formulation improves gradient flow
and mitigates vanishing gradients in the standard softmax.

The second source candidate, $f_s^2$, uses an averaged
softmax formulation that combines the standard and weighted
attention weights (eq. \ref{eq:fs3}):
\begin{equation}
  \label{eq:fs3}
  \alpha^{(3)}_{ij}
  =
  \frac{1}{2}\alpha^{(1)}_{ij}
  +
  \frac{1}{2}\alpha^{(2)}_{ij}
\end{equation}
Thus, $f_s^1$ is trained with $\alpha^{(1)}_{ij}$, while
$f_s^2$ is trained with $\alpha^{(3)}_{ij}$. 


We adopt $f_s^1$ and $f_s^2$ in DGASS because they provide
complementary initialization behaviour: $f_s^1$ gives sharp
discriminative focus, whereas $f_s^2$ balances sharpness
with gradient stability and yields a smoother alternative.
\noindent\textbf{Domain-gap estimation.}
We estimate domain gap severity from the prediction entropy
of the source model on unlabelled target samples.
The per-sample entropy for target sample $x_t^{(i)}$ is:
\begin{equation}
  \label{eq:persample_entropy}
  H\!\left(x_t^{(i)}\right)
  = -\sum_{k=1}^{C}
    p\!\left(y_t^k \,|\, x_t^{(i)}\right)
    \log p\!\left(y_t^k \,|\, x_t^{(i)}\right)
\end{equation}
where $p(y_t^k \,|\, x_t^{(i)}) = f_s^{(k)}(x_t^{(i)})$.
The overall uncertainty of a source model on the target
domain is quantified by the average entropy over all
$N_t$ target samples:
\begin{equation}
    \small
  \label{eq:entropy}
  \bar{H}(f_s;\,\mathcal{D}_t)
  = -\,\frac{1}{N_t}\sum_{i=1}^{N_t}
    \sum_{k=1}^{C}
    p\!\left(y_t^k \,|\, x_t^{(i)}\right)
    \log p\!\left(y_t^k \,|\, x_t^{(i)}\right)
\end{equation}
Low $\bar{H}$ signals confident source predictions,
indicating a small domain gap; high $\bar{H}$ signals
uncertainty, indicating a large gap.

\noindent\textbf{Entropy-guided selection.}
Based on the estimated gap severity, DGASS selects the
appropriate source variant:
%

%
\begin{equation}
\small
\label{eq:dgass}
f_s^{*}
=
\begin{cases}
\text{High-entropy variant}, & \bar{H} \geq \lambda\\[3pt]
\text{Low-entropy variant}, & \bar{H} < \lambda
\end{cases}
\end{equation}
\noindent\textit{Intuition.}
When the domain gap is large, a confident (low-entropy)
source model risks over-committing to incorrect pseudo-labels,
causing rapid error accumulation.
Selecting the higher-entropy variant instead encourages
cautious adaptation, providing room for gradual
self-correction.
Conversely, when the gap is small, the confident variant
provides reliable pseudo-labels that accelerate convergence.
This adaptive selection is evaluated against fixed source-selection
strategies in the experiments.

\noindent\textbf{Principled threshold selection.}
The threshold $\lambda$ is determined once per dataset
from unlabelled target predictions (no labels required).

Let $\mathcal{T}$ denote the set of all transfer tasks in the dataset,
and let $\mathcal{X}_{t}$ denote the unlabeled target samples for
task $t\in\mathcal{T}$. We define the automatic entropy threshold as:
\begin{equation}
\label{eq:lambda_auto}
\lambda_{\mathrm{}}
=
\operatorname{median}_{t\in\mathcal{T}}
\left\{
\min\left(
\bar{H}(f_s^{1};\mathcal{X}_{t}),
\bar{H}(f_s^{2};\mathcal{X}_{t})
\right)
\right\}
\end{equation}
The threshold $\lambda$ is computed from unlabeled target prediction entropy using Eq. \ref{eq:lambda_auto}. For each transfer task, we compute the average prediction entropy of both source models and retain the smaller entropy value for threshold estimation. The DGASS threshold $\lambda$ is set to $0.78$ for Office31, $1.10$ for Office-Home, and $0.30$ for DomainNet40. On DomainNet40, 11 of 12 tasks satisfy $\bar{H}\!\geq\!0.30$,
indicating predominantly high-gap transfer tasks.  This median-based rule provides an $\ell_1$-robust, distribution-free
operating point for separating relatively low-gap and high-gap transfers.
Additional motivation, robustness properties, and complete entropy statistics
are provided in the Supplementary.



\subsection{Training Objective}
\label{sec:method_training}

The student is optimised by four complementary losses.
The \textbf{Knowledge Distillation} loss aligns the student's
soft predictions with EMA teacher pseudo-labels:
\begin{equation}
  \label{eq:lkd}
  \mathcal{L}_{\mathrm{KD}}
  = -\,\mathbb{E}_{x_t \sim \mathcal{D}_t}
    \left[
      \sum_{k=1}^{C}
      \hat{y}_t^{(k)}
      \log f_t^{s(k)}(x_t)
    \right]
\end{equation}
where $\hat{y}_t$ are soft pseudo-labels generated by
$f_{\mathrm{ema}}$.
The \textbf{Classifier} loss sharpens decision boundaries
using hard pseudo-labels ($ \hat{\mathbf{y}}_t^{(k)}$) for confident predictions:
\begin{equation}
  \label{eq:lcl}
  \mathcal{L}_{\mathrm{CL}}
  = -\,\mathbb{E}_{x_t \sim \mathcal{D}_t}
    \left[\hat{\mathbf{y}}_t \log f_t^s(x_t)\right]
\end{equation}
%
The \textbf{Information Maximisation} loss balances
per-sample prediction confidence with marginal class
diversity.
%
Let
$\mathbf{p}(x_t)=\sigma(f_t(x_t))$, where $p_k(x_t)$ denotes the
$k$-th class probability and
$\bar{p}_k=\mathbb{E}_{x_t}[p_k(x_t)]$ denotes the empirical marginal
probability of class $k$. We define
\begin{equation}
\small
\label{eq:lim}
\mathcal{L}_{\mathrm{IM}}
=
-\mathbb{E}_{x_t}
\left[
\sum_{k=1}^{C}
p_k(x_t)\log p_k(x_t)
\right]
+
\sum_{k=1}^{C}
\bar{p}_k\log \bar{p}_k
\end{equation}
Minimizing the first term encourages confident per-sample predictions,
while minimizing the second term is equivalent to maximizing the
marginal entropy $H(\bar{\mathbf{p}})$, thereby discouraging class
collapse.
The complete adaptation objective is:
\begin{equation}
  \label{eq:ltotal}
  \mathcal{L}_{\mathrm{total}}
  = \mathcal{L}_{\mathrm{CAFC}}
  + \mathcal{L}_{\mathrm{IM}}
  + 0.3\,\mathcal{L}_{\mathrm{KD}}
  + \mathcal{L}_{\mathrm{CL}}
\end{equation}
%
The weight $0.3$ on $\mathcal{L}_{\mathrm{KD}}$ down-weights the
teacher signal early in adaptation when pseudo-labels are less reliable;
all other loss weights are $1.0$. The EMA teacher is updated after each
student step with momentum $m=0.999$
\begin{equation}
\label{eq:ema}
{\theta}_{\mathrm{ema}}
\leftarrow
m\,{\theta}_{\mathrm{ema}}
+
(1-m)\,{\theta}_{t}
\end{equation}

\subsection{Extension to Domain Generalization}
\label{sec:method_dg}

In the domain generalization (DG) setting, no target data
are available during training, making DGASS inapplicable.
We therefore apply $\mathcal{L}_{\mathrm{CAFC}}$ alone as a
lightweight regularizer during source-domain training. The
same counterfactual attention perturbations and CAFC loss are
computed on source-domain batches without requiring target
data. Since the ResNet backbones used here do not natively produce transformer-style attention maps, we append a lightweight self-attention adapter: the 512-dimensional
pooled feature is projected into $N=8$ latent tokens, processed
by a single self-attention layer, and pooled back to 512 dimensions
before classification. We evaluate CAFC under two protocols: a ResNet-18-from-scratch isolation setting and a standard DomainBed-style setting with ImageNet-pretrained ResNet-50. The former isolates the contribution of CAFC from ImageNet pretraining, while the latter enables direct comparison with standard DG evaluations.

\section{Experimental Setup}
\label{sec:exp_setup}

\noindent\textbf{Datasets.}
For SFDA, we evaluate on Office31, Office-Home, and
DomainNet40. Office31 contains three domains
(Amazon, Webcam, DSLR) and 31 classes; Office-Home
contains four domains (Artistic, ClipArt, Product, Real-World)
and 65 classes; DomainNet40 contains four domains
(Clipart, Painting, Real, Sketch) and 40 classes. For DG,
we evaluate on PACS and VLCS under the standard
leave-one-domain-out protocol.


\noindent\textbf{Implementation details.}
For SFDA, we use an ImageNet-pretrained ResNet-50
backbone followed by transformer attention layers, a
bottleneck layer, and a classifier. Source models are trained
with label-smoothed cross-entropy using $\beta=0.1$.
During adaptation, the DGASS-selected source hypothesis
$f_s^{*}$ initializes both the teacher and student models.
The EMA teacher uses momentum $m=0.999$. The student is
optimized using SGD with learning rate $1\times10^{-3}$,
momentum $0.9$, and weight decay $10^{-3}$. Images are
resized to $224\times224$, batch size is $64$, and adaptation
runs for $15$ epochs. The loss weights are fixed as
$1.0$ for CAFC, $1.0$ for information maximization, $0.3$
for knowledge distillation, and $1.0$ for cross-entropy.
The DGASS thresholds are set without target labels:
$0.78$ for Office31, $1.10$ for Office-Home, and $0.30$
for DomainNet40.
For DG, we use ResNet-18 trained from scratch as well as ImageNet
pretrained ResNet-50. A lightweight
self-attention adapter maps the 512-dimensional backbone
feature into $N=8$ latent tokens to compute the CAFC
regularization signal. The CAFC coefficient is fixed as
$\lambda_{\mathrm{CAFC}}=1.0$ for all DG objectives. All
experiments use the same image size, split protocol, and
three-seed evaluation. \textbf{Evaluation.}
We report top-1 accuracy. SFDA results are averaged over
all transfer tasks, while DG results are averaged over held-out
target domains. All reported results are mean over three
independent seeds; DG tables additionally report standard
deviation. 

\begin{table*}[t]
  \centering
  \scalebox{0.7}{
  \begin{tabular}{@{}lccccccccccccc@{}}
    \toprule
    Method
      & A{$\to$}C & A{$\to$}P & A{$\to$}R
      & C{$\to$}A & C{$\to$}P & C{$\to$}R
      & P{$\to$}A & P{$\to$}C & P{$\to$}R
      & R{$\to$}A & R{$\to$}C & R{$\to$}P
      & Avg. \\
    \midrule

    Source Only
      & 59.47 & 74.32 & 80.31 & 64.44 & 72.72 & 74.91
      & 69.67 & 58.10 & 82.49 & 76.10 & 61.40 & 83.67
      & 71.47 \\

    GSDA~\cite{hu2020unsupervised}
      & 61.30 & 76.10 & 79.40 & 65.40 & 73.30 & 74.30
      & 65.00 & 53.20 & 80.00 & 72.20 & 60.60 & 83.10
      & 70.30 \\

    SHOT~\cite{liang2020we}
      & 57.10 & 78.10 & 81.50 & 68.00 & 78.20 & 78.10
      & 67.40 & 54.90 & 82.20 & 73.30 & 58.80 & 84.30
      & 71.80 \\

    NRC~\cite{yang2021exploiting}
      & 57.70 & 80.30 & 82.00 & 68.10 & 79.80 & 78.60
      & 65.30 & 56.40 & 83.00 & 71.00 & 58.60 & 85.60
      & 72.20 \\

    AaD~\cite{yang2022attracting}
      & 59.30 & 79.30 & 82.10 & 68.90 & 79.80 & 79.50
      & 67.20 & 57.40 & 83.10 & 72.10 & 58.50 & 85.40
      & 72.70 \\

    CoWA-JMDS~\cite{lee2022confidence}
      & 56.90 & 78.40 & 81.00 & 69.10 & 80.00 & 79.90
      & 67.70 & 57.20 & 82.40 & 72.80 & 60.50 & 84.50
      & 72.50 \\

    EKS~\cite{rai2025label}
      & 58.95 & 79.49 & 82.86 & 70.62 & 80.22 & 81.48
      & 71.80 & 57.21 & 83.76 & 75.94 & 62.27 & 86.30
      & 74.24 \\

    TVT~\cite{yang2108tvt}
      & 67.10 & \textbf{83.50} & \textbf{87.30} & \textbf{77.40}
      & \underline{85.00} & \textbf{85.60}
      & 75.60 & 64.90 & 86.60 & 79.10 & 67.20 & 88.00
      & 78.90 \\

    TransDA~\cite{yang2021transformer}
      & \underline{67.50} & \underline{83.30} & 85.90 & 74.00
      & 83.80 & 84.40
      & \underline{77.00} & \textbf{68.00} & \textbf{87.00}
      & \textbf{80.50} & \textbf{69.90} & \textbf{90.00}
      & \underline{79.30} \\

    \midrule
    PLUE$^{\dagger}$~\cite{litrico2023guiding}
      & 49.10 & 73.50 & 78.20 & 62.90 & 73.50 & 74.50
      & 62.20 & 48.30 & 78.60 & 68.60 & 51.80 & 81.50
      & 66.90 \\

    C-SFDA~\cite{karim2023c}
      & 60.30 & 80.20 & 82.90 & 69.30 & 80.10 & 78.80
      & 67.30 & 58.10 & 83.40 & 73.60 & 61.30 & 86.30
      & 73.50 \\

    TPDS~\cite{tang2024source}
      & 59.30 & 80.30 & 82.10 & 70.60 & 79.40 & 80.90
      & 69.80 & 56.80 & 82.10 & 74.50 & 61.20 & 85.30
      & 73.50 \\

    DPC~\cite{xia2024discriminative}
      & 59.50 & 80.60 & 82.90 & 69.40 & 79.30 & 80.10
      & 67.30 & 57.20 & 83.70 & 73.10 & 58.90 & 84.90
      & 73.10 \\

    ASM~\cite{jing2024visually}
      & 56.90 & 79.10 & 82.90 & 69.50 & 79.60 & 79.70
      & 67.90 & 55.10 & 82.60 & 74.70 & 60.50 & 84.80
      & 72.80 \\

    CPD~\cite{zhou2024source}
      & 59.10 & 79.00 & 82.40 & 68.50 & 79.70 & 79.50
      & 67.90 & 57.90 & 82.80 & 73.80 & 61.20 & 84.60
      & 73.00 \\

    UCon~\cite{xu2025revisiting}
      & 61.50 & 80.50 & 82.10 & 69.30 & 80.80 & 78.70
      & 67.00 & 62.20 & 82.00 & 72.20 & 61.90 & 85.50
      & 73.60 \\

    \midrule

    \textbf{DGA-CAFC (Ours)}
      & \textbf{69.44} & 82.95 & \underline{86.27}
      & \underline{76.10} & \textbf{86.39} & \underline{84.58}
      & \textbf{78.24} & \underline{67.17} & \underline{86.89}
      & \underline{79.65} & \underline{69.48} & \underline{88.76}
      & \textbf{79.66} \\

    \bottomrule
  \end{tabular}%
  }
  \vspace{-0.8em}
  \caption{Classification accuracy (\%) on Office-Home under the
  source-free domain adaptation protocol.}
  \label{tab:officehome}
  \vspace{-0.5em}
\end{table*}

\begin{table*}[t]
\centering

\scalebox{0.7}{
\begin{tabular}{lccccccccccccc}
\toprule
Method
  & S{$\to$}C & S{$\to$}P & S{$\to$}R
  & C{$\to$}S & C{$\to$}P & C{$\to$}R
  & P{$\to$}S & P{$\to$}C & P{$\to$}R
  & R{$\to$}S & R{$\to$}C & R{$\to$}P
  & Avg. \\
\midrule
Source Only
  & 80.22 & 79.47 & 84.78
  & 76.12 & 73.82 & 84.23
  & 78.30 & 76.61 & 89.23
  & 73.08 & 79.45 & 84.43
  & 79.98 \\
SHOT~\cite{liang2020we}
  & 80.50 & 70.80 & 88.30
  & 75.50 & 74.00 & 88.40
  & 76.20 & 77.70 & 89.80
  & 72.80 & 79.40 & 75.40
  & 79.10 \\
EKS~\cite{rai2025label}
  & 81.88 & 81.44 & 88.77
  & 78.69 & 82.91 & \underline{91.82}
  & 78.56 & 83.62 & \textbf{92.05}
  & 77.30 & 83.54 & 84.37
  & 83.74 \\
TransDA~\cite{yang2021transformer}
  & \underline{85.97} & \underline{86.31} & 89.41
  & \underline{82.72} & \textbf{85.42} & \underline{91.82}
  & \underline{83.42} & \textbf{85.37} & 91.39
  & \underline{81.26} & \underline{85.84} & \underline{86.80}
  & \underline{86.31} \\
SSRT-B~\cite{sun2022safe}
  & 70.60 & 62.20 & 73.20
  & 59.80 & 60.20 & 75.80
  & 55.20 & 61.70 & 71.40
  & 58.90 & 69.90 & 66.00
  & 65.41 \\
\midrule
\textbf{DGA-CAFC (Ours)}
  &{\textbf{87.19}} &{\textbf{89.32}} &{\textbf{93.47}}
  &{\textbf{84.15}} &\underline{85.39} &{\textbf{92.13}}
  &{\textbf{84.60}} &\underline{85.31} &{\underline{91.56}}
  &{\textbf{81.98}} &{\textbf{87.94}} &{\textbf{86.91}}
  &{\textbf{87.49}} \\
\bottomrule
\end{tabular}%
}
\vspace{-0.8em}
\caption{Classification accuracy (\%) on DomainNet40 under the
source-free domain adaptation protocol.}
\label{tab:domainnet40}
\vspace{-1.5em}
\end{table*}

\begin{table}[b]
\vspace{-0.8em}
  \centering
  \resizebox{\columnwidth}{!}{%
  \begin{tabular}{@{}lccccccc@{}}
    \toprule
    Method
      & A{$\to$}D & A{$\to$}W & D{$\to$}A
      & D{$\to$}W & W{$\to$}A & W{$\to$}D & Avg. \\
    \midrule

    Source Only
      & 83.53 & 85.91 & 59.74 & 97.74 & 64.89 & 99.60 & 81.90 \\

    SHOT~\cite{liang2020we}
      & 94.00 & 90.10 & 74.70 & 98.40 & 74.30
      & \underline{99.90} & 88.60 \\

    3C-GAN~\cite{agarwal2022unsupervised}
      & 92.70 & 93.70 & 75.30 & 98.50 & 77.80 & 99.80 & 89.60 \\

    NRC~\cite{yang2021exploiting}
      & 96.00 & 90.80 & 75.30 & \underline{99.10} & 75.00
      & \textbf{100.00} & 89.40 \\

    AaD~\cite{yang2022attracting}
      & 96.40 & 92.10 & 75.00 & \underline{99.10} & 76.50
      & \textbf{100.00} & 89.90 \\

    CoWA-JMDS~\cite{lee2022confidence}
      & 94.40 & 95.20 & 76.20 & 98.50 & 77.60 & 99.80 & 90.30 \\

    CAN~\cite{kang2019contrastive}
      & 95.00 & 94.50 & \textbf{78.00} & \underline{99.10}
      & 77.00 & 99.80 & 90.60 \\

    EKS~\cite{rai2025label}
      & \textbf{98.10} & \underline{96.40} & 74.50 & 99.00
      & 75.10 & 99.80 & 90.50 \\

    TransDA~\cite{yang2021transformer}
      & 97.20 & 95.00 & 73.70 & \textbf{99.30}
      & \textbf{79.30} & 99.60 & 90.70 \\

    \midrule

    PLUE$^{\dagger}$~\cite{litrico2023guiding}
      & 89.20 & 88.40 & 72.80 & 97.10 & 69.60 & 97.90 & 85.80 \\

    C-SFDA~\cite{karim2023c}
      & 96.20 & 93.90 & 77.30 & 98.80 & 77.90 & 99.70 & 90.50 \\

    TPDS~\cite{tang2024source}
      & 97.10 & 94.50 & 75.70 & 98.70 & 75.50 & 99.80 & 90.20 \\

    DPC~\cite{xia2024discriminative}
      & 95.80 & 94.50 & 76.50 & 98.90 & 76.80
      & \textbf{100.00} & 90.50 \\

    ASM~\cite{jing2024visually}
      & 96.00 & 95.10 & 75.30 & 98.70 & 77.20
      & \textbf{100.00} & 90.40 \\

    CPD~\cite{zhou2024source}
      & 96.60 & 94.20 & 77.30 & 98.20 & 78.30
      & \textbf{100.00} & \underline{90.80} \\

    UCon~\cite{xu2025revisiting}
      & 94.80 & 95.40 & 77.10 & 98.90 & 77.10
      & \textbf{100.00} & 90.60 \\

    \midrule

    \textbf{DGA-CAFC (Ours)}
      & \underline{97.59} & \textbf{96.48} & \underline{77.78}
      & 99.00 & \underline{79.06} & \textbf{100.00}
      & \textbf{91.65} \\

    \bottomrule
  \end{tabular}%
  }
  \vspace{-0.8em}
  \caption{Classification accuracy (\%) on Office31 under the SFDA.}
  \label{tab:office31}
\end{table}

\section{Experimental Results}
\label{sec:results}

\begin{table}[b]
\centering
\tiny
\setlength{\tabcolsep}{2.2pt}
\renewcommand{\arraystretch}{0.78}
\resizebox{\columnwidth}{!}{%
\begin{tabular}{llccccc}
\toprule
Method & Backbone & Art & Cartoon & Photo & Sketch & Avg. \\
\midrule

ERM & R-18 
& 39.73$\pm$1.65 & 51.68$\pm$3.29 & 66.75$\pm$0.73
& 39.55$\pm$3.18 & 49.43$\pm$1.59 \\

ERM+CAFC & R-18 
& 39.99$\pm$1.85 & 54.76$\pm$2.99 & 65.49$\pm$2.77
& 41.88$\pm$10.12
& \textbf{50.53$\pm$1.70} \textbf{(+1.10)} \\

ERM & R-50 
& 80.55$\pm$2.57 & 76.07$\pm$2.95 & 97.72$\pm$0.10
& 62.14$\pm$6.76 & 79.12$\pm$1.49 \\

ERM+CAFC & R-50 
& 81.74$\pm$1.57 & 77.16$\pm$2.87 & 97.96$\pm$0.33
& 69.02$\pm$2.26
& \textbf{81.47$\pm$1.08} \textbf{(+2.35)} \\

\midrule

CORAL~\cite{sun2016deep} & R-18 
& 35.95$\pm$2.20 & 50.51$\pm$1.65 & 55.53$\pm$3.37
& 41.11$\pm$0.17 & 45.78$\pm$1.70 \\

CORAL+CAFC & R-18 
& 40.10$\pm$3.76 & 51.86$\pm$2.33 & 52.65$\pm$1.44
& 43.71$\pm$4.61
& \textbf{47.08$\pm$2.37} \textbf{(+1.30)} \\

CORAL & R-50 
& 80.21$\pm$1.29 & 77.39$\pm$1.85 & 95.67$\pm$0.71
& 72.73$\pm$2.10 & 81.50$\pm$0.50 \\

CORAL+CAFC & R-50 
& 81.85$\pm$3.87 & 75.60$\pm$0.91 & 96.26$\pm$0.87
& 75.36$\pm$1.78
& \textbf{82.27$\pm$0.43} \textbf{(+0.77)} \\

\midrule

Mixup~\cite{zhang2017mixup} & R-18
& 35.84$\pm$8.22 & 47.72$\pm$4.35 & 53.59$\pm$5.90
& 35.05$\pm$11.23 & 43.05$\pm$5.52 \\

Mixup+CAFC & R-18 
& 40.30$\pm$3.25 & 51.66$\pm$3.06 & 57.96$\pm$6.77
& 39.82$\pm$10.90
& \textbf{47.44$\pm$2.46} \textbf{(+4.39)} \\

Mixup & R-50 
& 86.25$\pm$1.09 & 79.36$\pm$0.15 & 96.17$\pm$1.51
& 72.74$\pm$3.96 & 83.63$\pm$1.31 \\

Mixup+CAFC & R-50
& 87.50$\pm$2.29 & 78.24$\pm$0.17 & 97.26$\pm$1.06
& 75.17$\pm$2.25
& \textbf{84.54$\pm$1.30} \textbf{(+0.91)} \\

\midrule

GroupDRO~\cite{sagawa2019distributionally} & R-18 
& 34.29$\pm$3.20 & 51.35$\pm$2.96 & 48.68$\pm$3.32
& 46.01$\pm$5.08 & 45.08$\pm$1.85 \\

GroupDRO+CAFC & R-18 
& 39.40$\pm$1.93 & 53.50$\pm$0.91 & 53.57$\pm$3.48
& 50.40$\pm$2.31
& \textbf{49.22$\pm$1.06} \textbf{(+4.14)} \\

GroupDRO & R-50 
& 80.13$\pm$0.73 & 74.39$\pm$1.43 & 96.57$\pm$0.43
& 74.65$\pm$7.02 & 81.43$\pm$1.94 \\

GroupDRO+CAFC & R-50 
& 80.78$\pm$4.46 & 77.50$\pm$3.59 & 96.03$\pm$0.73
& 77.46$\pm$2.74
& \textbf{82.94$\pm$1.93} \textbf{(+1.51)} \\

\midrule

VREx~\cite{krueger2021out} & R-18 
& 37.17$\pm$2.65 & 53.31$\pm$2.08 & 55.43$\pm$2.38
& 47.31$\pm$3.67
& \textbf{48.31$\pm$1.75} \\

VREx+CAFC & R-18 
& 41.57$\pm$4.17 & 49.23$\pm$2.50 & 53.79$\pm$5.28
& 45.29$\pm$10.36
& 47.47$\pm$3.83 (-0.84) \\

VREx & R-50 
& 78.19$\pm$3.37 & 74.69$\pm$4.49 & 95.47$\pm$0.30
& 70.75$\pm$8.65 & 79.77$\pm$2.86 \\

VREx+CAFC & R-50 
& 79.51$\pm$2.61 & 75.78$\pm$2.54 & 95.29$\pm$0.78
& 72.76$\pm$0.37
& \textbf{80.84$\pm$0.31} \textbf{(+1.07)} \\

\bottomrule
\end{tabular}}
\caption{DG results on PACS with ResNet-18-from-scratch and pretrained
ResNet-50. Parentheses denote gains over the baseline.}
\label{tab:pacs_combined_dg}
\end{table}

\begin{table}[t]
\centering
\tiny
\setlength{\tabcolsep}{2.2pt}
\renewcommand{\arraystretch}{0.78}
\resizebox{\columnwidth}{!}{%
\begin{tabular}{llccccc}
\toprule
Method & Backbone & Caltech101 & LabelMe & SUN09 & VOC2007 & Avg. \\
\midrule

ERM & R-18 
& 73.10$\pm$3.16 & 55.17$\pm$2.02 & 48.03$\pm$3.77
& 49.40$\pm$1.47 & 56.42$\pm$1.64 \\

ERM+CAFC & R-18 
& 76.09$\pm$3.23 & 55.46$\pm$3.80 & 47.86$\pm$3.04
& 48.46$\pm$4.87
& \textbf{56.97$\pm$1.84} \textbf{(+0.55)} \\

ERM & R-50 
& 98.04$\pm$0.68 & 65.75$\pm$0.98 & 74.01$\pm$0.80
& 71.36$\pm$3.33 & 77.29$\pm$0.41 \\

ERM+CAFC & R-50 
& 98.03$\pm$0.23 & 67.44$\pm$0.38 & 77.25$\pm$1.58
& 78.97$\pm$2.70
& \textbf{80.42$\pm$0.49} \textbf{(+3.13)} \\

\midrule

CORAL~\cite{sun2016deep} & R-18 
& 73.07$\pm$2.23 & 55.41$\pm$0.93 & 49.27$\pm$4.03
& 43.93$\pm$0.81 & 55.42$\pm$0.56 \\

CORAL+CAFC & R-18 
& 73.80$\pm$2.44 & 56.25$\pm$2.36 & 49.69$\pm$2.75
& 48.01$\pm$0.29
& \textbf{56.94$\pm$0.81} \textbf{(+1.52)} \\

CORAL & R-50 
& 98.26$\pm$0.57 & 65.13$\pm$0.52 & 75.10$\pm$0.67
& 72.38$\pm$1.82 & 77.72$\pm$0.11 \\

CORAL+CAFC & R-50 
& 98.53$\pm$0.41 & 66.20$\pm$0.96 & 74.61$\pm$2.67
& 76.18$\pm$1.01
& \textbf{78.88$\pm$0.68} \textbf{(+1.16)} \\

\midrule

Mixup~\cite{zhang2017mixup} & R-18 
& 71.14$\pm$1.42 & 57.67$\pm$3.31 & 50.77$\pm$3.49
& 47.01$\pm$1.81 & 56.65$\pm$0.89 \\

Mixup+CAFC & R-18 
& 75.57$\pm$1.22 & 57.67$\pm$1.20 & 51.52$\pm$2.25
& 47.63$\pm$2.41
& \textbf{58.10$\pm$0.36} \textbf{(+1.45)} \\

Mixup & R-50 
& 98.57$\pm$0.11 & 65.61$\pm$1.87 & 73.55$\pm$1.64
& 77.58$\pm$1.37 & 78.83$\pm$0.34 \\

Mixup+CAFC & R-50 
& 98.65$\pm$0.23 & 67.07$\pm$1.62 & 76.36$\pm$1.00
& 76.66$\pm$2.11
& \textbf{79.69$\pm$0.58} \textbf{(+0.86)} \\

\midrule

GroupDRO~\cite{sagawa2019distributionally} & R-18 
& 76.07$\pm$0.76 & 54.61$\pm$5.07 & 51.29$\pm$1.40
& 46.30$\pm$2.18 & 57.06$\pm$1.79 \\

GroupDRO+CAFC & R-18 
& 76.87$\pm$2.90 & 58.41$\pm$2.46 & 52.75$\pm$4.15
& 47.42$\pm$0.65
& \textbf{58.86$\pm$0.15} \textbf{(+1.80)} \\

GroupDRO & R-50 
& 98.52$\pm$0.51 & 65.04$\pm$0.35 & 75.50$\pm$1.09
& 70.02$\pm$1.12 & 77.27$\pm$0.29 \\

GroupDRO+CAFC & R-50 
& 98.56$\pm$0.25 & 66.03$\pm$2.00 & 75.50$\pm$0.90
& 74.93$\pm$1.75
& \textbf{78.76$\pm$0.58} \textbf{(+1.49)} \\

\midrule

MMD~\cite{li2024bridging} & R-18 
& 73.59$\pm$1.13 & 54.43$\pm$0.97 & 49.28$\pm$3.42
& 45.45$\pm$1.45 & 55.69$\pm$0.91 \\

MMD+CAFC & R-18 
& 76.40$\pm$2.00 & 55.36$\pm$1.79 & 51.93$\pm$5.49
& 46.50$\pm$1.91
& \textbf{57.55$\pm$1.62} \textbf{(+1.86)} \\

MMD & R-50 
& 97.79$\pm$0.82 & 65.51$\pm$0.23 & 74.83$\pm$3.11
& 72.37$\pm$1.17 & 77.63$\pm$0.73 \\

MMD+CAFC & R-50 
& 98.07$\pm$0.82 & 67.75$\pm$0.60 & 75.25$\pm$1.16
& 76.00$\pm$0.94
& \textbf{79.27$\pm$0.60} \textbf{(+1.64)} \\

\bottomrule
\end{tabular}}
\vspace{-1.6em}
\caption{DG results on VLCS with ResNet-18-from-scratch and pretrained
ResNet-50. Parentheses denote gains over the baseline.}
\label{tab:vlcs_combined_dg}
\vspace{-1.5em}
\end{table}
We evaluate our method in two complementary settings:
\textit{source-free domain adaptation} (SFDA), where unlabeled
target data are available at adaptation time, and
\textit{domain generalization} (DG), where target data are
entirely withheld during training.
SFDA experiments evaluate the complete framework (CAFC + DGASS),
while DG experiments isolate CAFC as a plug-in attention-drift
regularizer applied to standard DG objectives without access to
target data.
\noindent \textbf{Source-Free Domain Adaptation Results}\\
\noindent\textbf{DomainNet40.}
\textbf{Table~\ref{tab:domainnet40}} reports classification accuracy on
DomainNet40, a challenging large-scale benchmark comprising four
domains (Clipart, Painting, Real, Sketch) and 40 classes.
Our method achieves an average accuracy of \textbf{{87.49\%}}, improving
over the strongest baseline TransDA~\cite{yang2021transformer}
by \textbf{+1.18 \%}.
Consistent gains are observed across the majority of transfer
pairs; notably, $\mathrm{S{\to}R}$ improves by +4.06 \% and
$\mathrm{P{\to}S}$ by +1.18 \%, reflecting the benefit of
attention-drift regularization under large style shifts.
We note that the gain on DomainNet40 is driven primarily by the
CAFC module
,
the uniformly high domain gap of this benchmark causes DGASS to
correctly default to the high-entropy source variant across
11 of 12 tasks.

\textbf{\noindent\textbf{Office31.}}
\textbf{Table~\ref{tab:office31}} lists results on Office31, which
comprises three domains (Amazon, Webcam, DSLR) and 31 classes.
Our method attains an average accuracy of {91.65\%},
surpassing all compared methods.
Relative to the recent SFDA baselines, our method outperforms NRC~\cite{yang2021exploiting}
by +2.30 \%, AaD~\cite{yang2022attracting} by +1.80 \%, and
CoWA-JMDS~\cite{lee2022confidence} by +1.37 \%.
Task-level analysis confirms our method achieves best or
second-best accuracy on five of six transfer tasks,
including the most competitive pair $\mathrm{D{\to}A}$
where accurate adaptation of dense appearance differences
benefits from DGASS-driven domain-gap-aware initialization.

\noindent\textbf{Office-Home.}
\textbf{Table~\ref{tab:officehome}} presents results on the more
challenging Office-Home benchmark (four domains, 65 classes,
12 transfer tasks).
Our method achieves the highest average accuracy of {79.66\%},
narrowly surpassing TransDA (79.30\%) and TVT~\cite{yang2108tvt}
(78.90\%) while outperforming recent SFDA specific methods
(NRC: 72.20\%, AaD: 72.70\%, CoWA-JMDS: 72.50\%).
Task-level inspection reveals our method ranks first on three tasks
($\mathrm{A{\to}C}$, $\mathrm{C{\to}P}$, $\mathrm{P{\to}A}$)
and second on seven further tasks, yielding consistent
improvement across diverse cross-domain appearance shifts.

\begin{table}[t]
\centering

\scalebox{0.7}{
\begin{tabular}{ccccc}
\toprule
CAFC & DGASS & Avg. & $\Delta$ & Note \\
\midrule
           &            & 90.64 & --      & Baseline \\
           & \checkmark & 91.11 & +0.47   & DGASS only \\
\checkmark &            & 91.01 & +0.37   & CAFC only \\
\checkmark & \checkmark &{91.65} & +1.01 & Full (\textit{interaction +0.17}) \\
\bottomrule
\end{tabular}
}
\vspace{-0.8em}
\caption{\small Ablation study of CAFC and DGASS on Office31.
The baseline uses neither module. $\Delta$ denotes the absolute
improvement over the baseline.}
  \vspace{-0.7em}
  \label{cafc_dgass_ablation}
\end{table}

\begin{table}[t]
\centering
\scalebox{0.55}{
\begin{tabular}{llcc}
\toprule
Method & Backbone & Office31 Avg. & Office-Home Avg. \\
\midrule
SHOT & ResNet-50 & 88.60 & 71.80 \\
 & Our ResNet-50+Attn & 87.67 & 70.53 \\
\midrule
CoWA-JMDS & ResNet-50 & 90.30 & 72.50 \\
 & Our ResNet-50+Attn & 87.40 & 71.49 \\
\midrule
NRC & ResNet-50 & 89.40 & 72.20 \\
 & Our ResNet-50+Attn & 82.31 & 64.06 \\
\midrule
\textbf{DGA-CAFC}
& \textbf{Our ResNet-50+Attn}
& \textbf{91.65} & \textbf{79.66} \\
\bottomrule
\end{tabular}}
\vspace{-0.7em}
\caption{Backbone-confound check: SFDA baselines on original ResNet-50 and matched ResNet-50+attention backbones.}
\label{tab:backbone_confound}
\vspace{-1.7em}
\end{table}


\noindent \textbf{Domain Generalization Results}
To evaluate whether CAFC generalizes beyond the source-free
adaptation setting, we integrate it as a plug-in regularizer
into various representative DG objectives on PACS and VLCS.
Since DG provides no target data, DGASS is not applicable;
these experiments therefore isolate the contribution of CAFC.
We report both the original ResNet-18-from-scratch isolation
setting and the standard DomainBed-style ImageNet-pretrained
ResNet-50 setting. Results are reported as mean $\pm$ standard
deviation over three seeds.

\noindent\textbf{PACS.}
\textbf{Table~\ref{tab:pacs_combined_dg}} reports results under both
protocols on PACS. In the ResNet-18 isolation setting, CAFC improves
four of five objectives, with particularly strong gains for
Mixup (+4.39\%) and GroupDRO (+4.14\%), while also reducing
seed variance for both objectives. More importantly, under the
standard ImageNet-pretrained ResNet-50 protocol, CAFC improves
all five objectives. This includes VREx, for which the
$-0.84 \%$ degradation observed with ResNet-18 is reversed to
a +1.07\% gain under ResNet-50. Together, these results show
that the benefit of CAFC persists under the standard pretrained
DG regime and that the isolated negative VREx case does not
carry over to this setting.
\noindent\textbf{VLCS.}
\textbf{Table~\ref{tab:vlcs_combined_dg}} shows a consistent benefit
across both protocols on VLCS dataset. In the ResNet-18 isolation setting,
CAFC improves all five objectives and reduces seed variance
for several of them, including Mixup and GroupDRO. Under the
standard ImageNet-pretrained ResNet-50 protocol, CAFC again
improves all five objectives, with the largest gains observed
for ERM (+3.13\%), MMD (+1.64\%), and GroupDRO (+1.49\%).
This consistency across objectives and backbone settings further
supports the robustness of CAFC as a DG regularizer.
Overall, the DomainBed-style ResNet-50 results strengthen
the DG evidence by showing improvements across all
10 objective$\times$dataset combinations under the standard
pretrained protocol. Together with the ResNet-18 isolation
results, this indicates that CAFC provides a broadly applicable
attention-stability regularization signal across different DG
objectives and backbone settings.
\section{Analysis}
\label{sec:analysis}


\noindent \textbf{Contribution of CAFC and DGASS Modules.} 
\textbf{Table~\ref{cafc_dgass_ablation}} reports the Office31 ablation study.
The baseline achieves 90.64\%, while adding DGASS alone improves
accuracy to 91.11\% (+0.47 \%) and adding CAFC alone improves it
to 91.01\% (+0.37 \%). The full DGA-CAFC framework reaches
91.65\%, giving a total gain of +1.01 \%. The positive interaction
term (+0.17 \%) shows that DGASS and CAFC are complementary:
domain-gap-aware initialization improves pseudo-label quality,
making CAFC regularization more effective during adaptation. This synergy is mechanistically intuitive, a source model
initialized with an appropriate confidence level for the
target domain produces pseudo-labels of higher quality,
which in turn makes CAFC consistency signal more
discriminative for adaptation.

\noindent \textbf{Backbone-Confound Analysis.} 
To isolate backbone effects from the algorithmic contribution, we re-run
SHOT, CoWA-JMDS, and NRC on the same ResNet-50+attention backbone used by
DGA-CAFC, using each method's official code and published hyperparameters.
Only the same lightweight self-attention module is added, with no other
architectural or training changes. Compared with the original ResNet-50
results in \textbf{Table~\ref{tab:backbone_confound}}, adding attention alone degrades
all three baselines, most severely NRC ($-7.04$\% on Office31 and $-8.14$\%
on Office-Home). This is consistent with evidence that transformer relevance
maps can rely on spurious background regions under distribution shift, while
regularizing such signals can improve out-of-distribution
robustness~\cite{chefer2022optimizing}. Thus, attention capacity alone is
insufficient; its benefit depends on regularizing the resulting
attention-conditioned features. On the identical backbone, DGA-CAFC
substantially outperforms all re-run baselines on both datasets, showing that
its gains cannot be attributed to the stronger backbone alone.

\noindent \textbf{Ablation of loss components.} 
\textbf{Table~\ref{tab:loss_component_ablation}} reports the contribution
of each adaptation loss while keeping CAFC and DGASS fixed.
Removing any loss degrades performance, confirming that IM, KD,
and CL provide complementary supervision during target adaptation.
The full objective achieves the best Office31 accuracy of
$91.65\%$.

\begin{table}[]
\centering
\scalebox{0.52}{
\begin{tabular}{lccccc c}
\toprule
Method & IM & KD & CL & CAFC & DGASS & Office-31 \\
\midrule
w/o IM        & -- & \checkmark & \checkmark & \checkmark & \checkmark & 90.28 \\
w/o KD        & \checkmark & -- & \checkmark & \checkmark & \checkmark & 90.41 \\
w/o CL        & \checkmark & \checkmark & -- & \checkmark & \checkmark & 90.58 \\
Full (DGA-CAFC)        & \checkmark & \checkmark & \checkmark & \checkmark & \checkmark & \textbf{91.65} \\
\bottomrule
\end{tabular}}
\vspace{-1em}
\caption{\small Loss-component ablation on Office31. CAFC and DGASS are kept fixed, while each of the losses are removed.}
\vspace{-0.8em}
\label{tab:loss_component_ablation}
\end{table}
\begin{table}[t]
\centering
\scalebox{0.6}{
\begin{tabular}{lcccccc}
\toprule
$\lambda$ & 0.0 & 0.2 & 0.78 & 0.8 & 1.6 & 1.7 \\
\midrule
Avg. & 90.01 & 90.01 &{91.65} & 91.53 & 91.53 & 91.46 \\
\bottomrule
\end{tabular}}
\vspace{-0.8em}
\caption{\small Sensitivity of DGA-CAFC to the DGASS threshold
$\lambda$ on Office31. Accuracy remains stable over a broad
range around the selected value.}
\label{tab:lambda}
\vspace{-1em}
\end{table}
\begin{table}[t]
\centering
\setlength{\tabcolsep}{3pt}
\resizebox{\columnwidth}{!}{%
\begin{tabular}{lcccccccccccccc}
\toprule
Method & A{$\to$}C & A{$\to$}P & A{$\to$}R & C{$\to$}A & C{$\to$}P & C{$\to$}R & P{$\to$}A & P{$\to$}C & P{$\to$}R & R{$\to$}A & R{$\to$}P & R{$\to$}C & Avg. \\
\midrule
TransDA~\cite{yang2021transformer}
  & .2808 & .2123 & .1867 & .1798 & .2007 & .2043
  & .1833 & .2957 & .2101 & .1611 & .2885 & .1955
  & .2166 \\
\textbf{DGA-CAFC}
  &{.2237} &{.0403} &{.0359} &{.1519} &{.1772} &{.0384}
  &{.1253} &{.2459} &{.1906} &{.0191} &{.2731} &{.0198}
  &{.1284} \\
\bottomrule
\end{tabular}}
\vspace{-0.8em}
\caption{\small Expected Calibration Error on Office-Home.}
\label{tab:ece}
\vspace{-1.8em}
\end{table}

\noindent\textbf{Adaptive selection outperforms fixed strategies.}
\textbf{Table~\ref{tab:ablation_selection}} compares DGASS with fixed
low- and high-entropy source selection. On Office31, DGASS
achieves 91.65\%, compared with 91.40\% and 91.00\% for
fixed-low and fixed-high selection, respectively. The advantage
is larger on Office-Home, where DGASS reaches 79.66\% versus
77.43\% and 74.90\% for fixed-low and fixed-high selection.
On DomainNet40, where 11 of 12 transfer tasks are identified
as high-gap, DGASS achieves 87.49\% compared with 85.89\%
under fixed-low selection. These results show that the preferred
source initialization depends on domain-gap severity rather than
a single fixed choice. Full per-task selections and entropy values
are provided in the Supplementary.
\begin{table}[t]
\centering
\scalebox{0.6}{
\begin{tabular}{lccc}
\toprule
Dataset & Fixed low & Fixed high & DGASS \\
\midrule
Office31
& 91.40 & 91.00 & \textbf{91.65} \\
Office-Home
& 77.43 & 74.90 & \textbf{79.66} \\
DomainNet40
& 85.89 & 86.23 & \textbf{87.49} \\
\bottomrule
\end{tabular}}
\vspace{-0.7em}
\caption{Fixed source-model selection vs. adaptive DGASS.
}
\label{tab:ablation_selection}
\end{table}
\noindent\textbf{Principled threshold selection.}
The threshold $\lambda$ is selected by a one-time unlabeled
preliminary sweep over target predictions using the
auto-$\lambda$ rule: $\lambda$ is set as the median of
per-task average entropy observed across the target domain
under either source variant.

\noindent \textbf{Backbone Generalization to Transformers.} 
To validate the proposed approach beyond the hybrid
ResNet-50+attention design, we evaluate native transformer
backbones in both DG and SFDA (\textbf{Table \ref{tab:transformer_validation}}). In DG on VLCS, CAFC improves
ERM with Swin-T by +1.59\% and with ViT-B/16 by +1.46\%.
In SFDA on Office31, DGA-CAFC with ViT-B/16 reaches 93.43\%,
outperforming SHOT-ViT (91.40\%) by +2.03\%.
Together, these ablation results show that the proposed method remains
effective with native transformer backbones across both DG and
SFDA, and is not tied to the ResNet-50+attention architecture.
\begin{table}[t]
\vspace{-1.5em}
\centering
\begin{tabular}{@{}c@{\hspace{8mm}}c@{}}
\begin{minipage}[t]{0.46\linewidth}
\vspace{0pt}
\centering
\textbf{DG: VLCS}
\vspace{1mm}
\scalebox{0.75}{
\begin{tabular}{lccc}
\toprule
Backbone & ERM & +CAFC & Gain \\
\midrule
ResNet-50 & 77.29 & 80.42 & \textbf{+3.13} \\
\midrule
Swin-T    & 79.98 & 81.57 & \textbf{+1.59} \\
\midrule
ViT-B/16  & 79.12 & 80.58 & \textbf{+1.46} \\
\bottomrule
\end{tabular}
}
\end{minipage}
\hfill
&

\begin{minipage}[t]{0.46\linewidth}
\vspace{0pt}
\centering
\textbf{SFDA: Office31}
\vspace{1mm}
\scalebox{0.6}{
\begin{tabular}{llc}
\toprule
Method & Backbone & Avg. \\
\midrule
SHOT & ResNet-50 & 88.60 \\
SHOT & ResNet-50+Attn & 87.67 \\
SHOT & ViT-B/16 & 91.40 \\
\midrule
DGA-CAFC & ResNet-50+Attn & 91.65 \\
\textbf{DGA-CAFC} & \textbf{ViT-B/16} & \textbf{93.43} \\
\bottomrule
\end{tabular}
}
\end{minipage}
\end{tabular}
\vspace{-1.2em}
\caption{Native-transformer validation in DG and SFDA.}
\label{tab:transformer_validation}
\vspace{-2em}
\end{table}


\noindent \textbf{Sensitivity to the DGASS Threshold $\lambda$.} 
\textbf{Table~\ref{tab:lambda}} reports DGA-CAFC performance across
different values of $\lambda$ on Office31.
Peak accuracy is achieved at $\lambda\!=\!0.78$, which
closely matches the $\lambda$ value of $0.78$ derived
from the median rule, confirming
that the data-driven threshold selection recovers the
optimal operating point without label access.
Performance remains within $0.19$ \% for
$\lambda \in [0.78, 1.7]$, indicating robustness to the
precise threshold value once it lies within the appropriate
entropy range.
At extreme values ($\lambda\!=\!0.0$, $0.2$), all tasks
are classified as high-gap and selection collapses to a
fixed strategy, causing performance to drop.



\noindent \textbf{Calibration Analysis.} 
\textbf{Table~\ref{tab:ece}} reports Expected Calibration Error (ECE)
on Office-Home, measuring how well predicted confidence aligns
with accuracy.
DGA-CAFC reduces ECE from 0.2166 (TransDA) to
0.1284.
Improved calibration benefits SFDA by making the entropy signal used
by DGASS more reliable and by reducing the influence of over-confident pseudo-labels during self-training.

\noindent \textbf{Transferability of CAFC to DG.} 
The DG results in \textbf{Tables, ~\ref{tab:pacs_combined_dg}}\textbf{ and
\ref{tab:vlcs_combined_dg}} show that CAFC transfers beyond SFDA.
Across interpolation-based, robust-optimization, feature-alignment,
and ERM objectives, CAFC improves all 10 objective$\times$dataset
combinations under the standard pretrained ResNet-50 protocol.
This suggests that
CAFC addresses a common source of domain-shift sensitivity rather than
acting as a correction for a single training objective. \textbf{Statistical significance.}
Paired seed-level tests confirm significant gains on PACS and VLCS, both $p<0.001$; full
statistics are provided in the Supplementary. \\
\noindent\textbf{Direct attention-drift validation.}
On PACS, CAFC reduces the mean in-/out-of-domain attention-map
$\ell_1$ divergence from 0.0357 to 0.0181, providing direct evidence
that CAFC suppresses attention drift under domain shift.
Full per-domain results are provided in the Supplementary.
\noindent\textbf{Variance reduction in several objectives.}
Beyond mean accuracy gains, CAFC reduces seed variance for selected DG
objectives. In the ResNet-18 isolation setting, the standard deviation
decreases for GroupDRO from $\pm$1.85 to $\pm$1.06 and for Mixup from
$\pm$5.52 to $\pm$2.46 on PACS; on VLCS, it decreases for GroupDRO from
$\pm$1.79 to $\pm$0.15 and for Mixup from $\pm$0.89 to $\pm$0.36.
These reductions suggest improved training stability for these objectives,
while full seed-level statistics are provided in the Supplementary.


\begin{figure}[!t]
\vspace{-0.4em}
    \centering
    \includegraphics[width=0.24\linewidth]{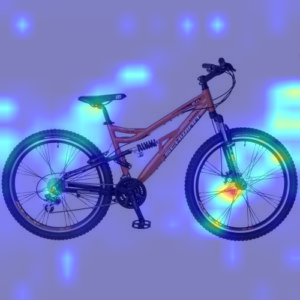}
    \includegraphics[width=0.24\linewidth]{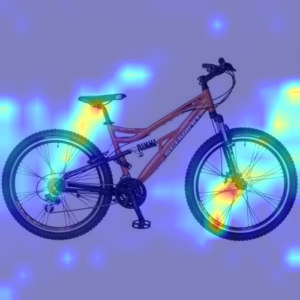}
    \vspace{-0.8em}
    \caption{\small Grad-CAM visualization for Office31 D→A. 
    }
    \label{fig:gradcam_dga_cafc}
    \vspace{-0.8em}
\end{figure}
\begin{figure}[!t]
    \centering
    \includegraphics[width=0.24\linewidth]{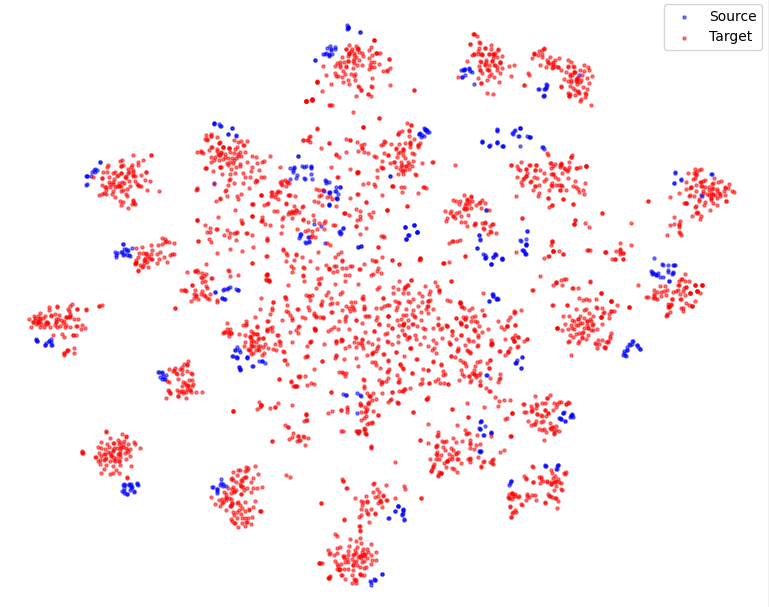}
    \includegraphics[width=0.24\linewidth]{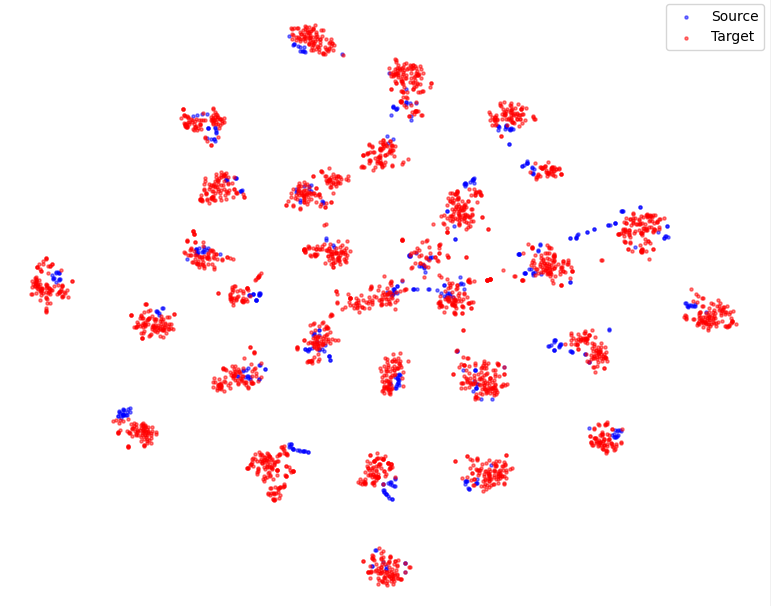}
    \vspace{-0.8em}
    \caption{\small t-SNE visualization for Office31 D→A. 
    }
    \label{fig:tsne_office_DA}
    \vspace{-1.7em}
\end{figure}

\noindent \textbf{Qualitative Analysis.} 
\textbf{Fig.~\ref{fig:gradcam_dga_cafc}} shows that CAFC broadens attention
beyond narrowly concentrated regions and shifts focus toward more
complete object structure on Office31 D$\to$A. This qualitatively
supports reduced reliance on domain-specific cues.
\textbf{Fig.~\ref{fig:tsne_office_DA}} shows that DGA-CAFC produces more
compact target clusters and clearer inter-class separation than the
source-only model, consistent with improved feature organization under
domain shift.
\vspace{-0.5em}
\section{Conclusion}
\label{sec:conclusion}
We presented DGA-CAFC, a source-free domain adaptation framework
designed to address two key limitations: attention drift under domain
shift and domain-gap agnostic source initialization. CAFC enforces
feature consistency under attention perturbations,
while DGASS selects a source model variant using target prediction
entropy to better match the estimated domain gap. Across Office31,
Office-Home, and DomainNet40, DGA-CAFC consistently improves SFDA
performance and calibration, with complementary gains from both
modules. CAFC also transfers to the DG setting, improving
standard DG objectives on PACS and VLCS without target data.
Results suggest that attention-stability regularization is a
useful principle for domain-robust recognition. 
\section*{Acknowledgment}
Shivangi Rai gratefully acknowledges the financial support provided through the Visvesvaraya PhD Fellowship by the Ministry of Electronics and Information Technology (MeitY), Government of India.
{
    \small
    \bibliographystyle{ieeenat_fullname}
    \bibliography{main}
}

\clearpage

\clearpage
\setcounter{page}{1}
\setcounter{section}{0}
\setcounter{table}{0}
\setcounter{figure}{0}
\setcounter{equation}{0}
\setcounter{algorithm}{0}

\renewcommand{\theHsection}{sup.\arabic{section}}
\renewcommand{\theHtable}{sup.\arabic{table}}
\renewcommand{\theHfigure}{sup.\arabic{figure}}
\renewcommand{\theHequation}{sup.\arabic{equation}}
\providecommand{\theHalgorithm}{\arabic{algorithm}}
\renewcommand{\theHalgorithm}{sup.\arabic{algorithm}}
\hypersetup{pageanchor=false}
\pagestyle{plain}

\twocolumn[{%
  \vspace*{0.25in}
  \begin{center}
    {\large\bfseries
    Supplementary Material of Attention Drift Regularization with Adaptive Initialization for Domain-Robust Visual Recognition\par}
  \end{center}
  \vspace{0.73in}
}]
\thispagestyle{plain}

\section{Quantification of the Attention Drift}

To directly quantify attention drift, we measure the divergence between learned attention distributions on the training domains and each unseen held-out domain in the DG setting. For each attention head, we compute the mean learned attention map over the training domains, $\bar{A}_{\mathrm{in}}$, and over the corresponding held-out domain, $\bar{A}_{\mathrm{out}}$. Attention drift is then defined as

\begin{equation}
D_{\mathrm{attn}}
=
\frac{1}{H}\sum_{h=1}^{H}
\frac{1}{N}\sum_{i=1}^{N}
\left\|
\bar{A}_{\mathrm{in}}^{(h)}[i,:]
-
\bar{A}_{\mathrm{out}}^{(h)}[i,:]
\right\|_1 
\end{equation}

Since each attention row is a probability distribution,
$D_{\mathrm{attn}}\in[0,2]$, with lower values indicating more stable attention across domain shift.

We compare ERM and ERM+CAFC using otherwise identical training and the same training-domain validation-based checkpoint selection. The analysis is performed on PACS for all four leave-one-domain-out settings (Art, Cartoon, Photo, and Sketch held out in turn), using seed 0. As shown in Table \ref{tab:attention_drift_pacs}, CAFC reduces attention divergence for every held-out domain. Averaged across the four targets, $D_{\mathrm{attn}}$ decreases from 0.0357 with ERM to 0.0181 with ERM+CAFC. This provides direct quantitative evidence that CAFC yields more domain-stable learned attention under distribution shift.

\begin{table}[h]
\centering
\resizebox{\linewidth}{!}{
\begin{tabular}{lcc}
\toprule
Held-out domain & ERM (no CAFC) & ERM+CAFC \\
\midrule
Art     & 0.0282 & \textbf{0.0144} \\
Cartoon & 0.0367 & \textbf{0.0172} \\
Photo   & 0.0307 & \textbf{0.0145} \\
Sketch  & 0.0471 & \textbf{0.0262} \\
\midrule
Mean    & 0.0357 & \textbf{0.0181} \\
\bottomrule
\end{tabular}}
\caption{Direct quantification of attention drift on PACS using in-domain vs.\ held-out-domain attention-map divergence. Lower $D_{\mathrm{attn}}$ indicates more stable attention across domain shift.}
\label{tab:attention_drift_pacs}
\end{table}
\section{Objective-Direction Analysis: CAL \cite{rao2021counterfactual} vs. CAFC}

Although CAFC uses counterfactual-inspired attention perturbations, its
objective is opposite to supervised counterfactual-attention methods such
as CAL. To test whether this distinction is empirically meaningful, we
compare CAFC with an attention-sensitivity variant that maximizes, rather
than minimizes, the learned-vs-counterfactual feature discrepancy. As shown
in Table~\ref{tab:cal_objective_ablation}, maximizing this discrepancy
reduces Office31 accuracy to $86.52\%$, below the baseline SFDA result of
$90.64\%$. In contrast, CAFC improves the baseline to $91.01\%$, and the
full DGA-CAFC framework reaches $91.65\%$. This confirms that the benefit
does not come merely from introducing counterfactual perturbations, but from
using them as a feature-stability regularizer under domain shift.

\begin{table}[t]
\centering
\scriptsize
\setlength{\tabcolsep}{3pt}
\renewcommand{\arraystretch}{0.90}
\resizebox{\columnwidth}{!}{%
\begin{tabular}{llc}
\toprule
Comparison & Method / Variant & Office31 Avg. \\
\midrule
\multirow{4}{*}{CAL vs. CAFC}
& Baseline SFDA & 90.64 \\
& Attention-sensitivity: maximize feature gap & 86.52 \\
& CAFC: minimize feature gap & 91.01 \\
& \textbf{DGA-CAFC: CAFC+DGASS} & \textbf{91.65} \\
\bottomrule
\end{tabular}}
\caption{Objective-direction ablation distinguishing CAFC from CAL.}
\label{tab:cal_objective_ablation}
\end{table}
\section{Detailed explanation of Self-labeling Losses}
Self-labeling losses leverage the model’s own outputs to
provide supervision on unlabeled target domain data. The
combination of these losses, knowledge distillation for soft
alignment, information maximization for confident and di-
verse predictions, and cross-entropy with hard pseudo-
labels helps to effectively adapt the model to new data dis-
tributions, enhancing its generalization capabilities.

\subsection{Knowledge Distillation Loss ($L_{KD}$)}

The Knowledge Distillation loss guides the student model to mimic the output distribution of a teacher model. It is defined as the cross-entropy between the soft predictions of the student model \( f_t^s(x_t) \) and the pseudo-labels \( \hat{y}_t \) generated by the teacher model:

\begin{equation}
L_{KD} = - \mathbb{E}_{x_t \sim D_t} \sum_{k=1}^C \hat{y}_t^{(k)} \log f_t^{s(k)}(x_t)
\end{equation}

Here, \( K \) denotes the number of classes, \( x_t \) represents the target sample, and \( \hat{y}_t^{(k)} \) is the teacher prediction for class \( k \). This loss encourages the student to align its class probability predictions with those of the teacher, thus transferring knowledge effectively.

\subsection{Information Maximization Loss ($L_{IM}$)}

The Information Maximization loss is designed to enhance the confidence and diversity of the student model's predictions. It combines an entropy minimization term and a marginal entropy maximization term:


\begin{multline}
L_{IM} = - \mathbb{E}_{x_t \sim D_t} \left[ 
    \sum_{k=1}^C \sigma \left( f_t^{s(k)}(x_t) \right) 
    \log \sigma \left( f_t^{s(k)}(x_t) \right) \right] \\
    + \sum_{k=1}^K \bar{p}_k \log \bar{p}_k
\end{multline}

In this equation, \( \sigma \) denotes the softmax function applied to the student model's output logits \( f_t^s(x_t) \), and \( \hat{p}_k \) represents the estimated marginal class distribution across the target domain. The first term encourages low-entropy (confident) predictions for individual samples, while the second term promotes high entropy over the marginal class distribution to avoid collapsed predictions that favor only a few classes. Together, these terms balance confidence and diversity, improving the robustness of the model on unlabeled data.

\subsection{Cross-Entropy Loss ($L_{CL}$)}
In cases where the source data distribution is similar to the target, we want to leverage the learning through pseudo label prediction as it does not contribute to the negative transfer.
The Cross-Entropy loss with hard pseudo labels uses the predicted class labels \( \hat{y}_t \) obtained by taking the argmax of the student’s output probabilities as the ground truth for the unlabeled samples:

\begin{equation}
L_{CL} = - \mathbb{E}_{x_t \sim D_t} \left[ \hat{y}_t \log f_t^s(x_t) \right]
\end{equation}
This loss treats the model's confident predictions as hard labels to supervise itself, promoting consistency in the decision boundaries. It is particularly useful when the model’s predictions are reliable enough to serve as ground truth.


\section{Necessity of Row-Normalisation for Counterfactual
Attention Maps}
\label{supp:norm}

We establish that row-wise $\ell_1$-normalisation of the
counterfactual attention maps $\mathbf{A}_{\mathrm{unif}}$,
$\mathbf{A}_{\mathrm{rand}}$, and $\mathbf{A}_{\mathrm{rev}}$
is necessary for $\mathcal{L}_{\mathrm{CAFC}}$ to measure
directional feature consistency rather than a scale artifact.\\
\textbf{Row-sum analysis of unnormalised maps.}
Let $N$ denote the sequence length.
The learned attention map $\mathbf{A}_L =
\mathrm{softmax}(\mathbf{Q}\mathbf{K}^\top/\sqrt{d})$
satisfies $\sum_j \mathbf{A}_L[i,j] = 1$ for all rows $i$
by construction.
The three unnormalised counterfactual maps have row sums:
\begin{align}
  \sum_j \mathbf{A}_{\mathrm{unif}}[i,j]
    &= N,
    \label{eq:sum_unif}\\[4pt]
  \sum_j \mathbf{A}_{\mathrm{rand}}[i,j]
    &\sim \mathcal{U}(0,\,N),
    \quad \mathbb{E}\bigl[\,\cdot\,\bigr] = N/2,
    \label{eq:sum_rand}\\[4pt]
  \sum_j \mathbf{A}_{\mathrm{rev}}[i,j]
    &= \sum_j (1 - \mathbf{A}_L[i,j])
     = N - 1.
    \label{eq:sum_rev}
\end{align}
None of these equals $1$, so the three maps are not valid
probability distributions over tokens.

\paragraph{Scale contamination without normalisation.}
Define the unnormalised aggregate:
\begin{equation}
  \tilde{\mathbf{A}}_{cf}
  = \frac{
      \mathbf{A}_{\mathrm{unif}}
    + \mathbf{A}_{\mathrm{rand}}
    + \mathbf{A}_{\mathrm{rev}}
    }{3},
\end{equation}
with per-row sum
$s_i \approx (N + N/2 + N - 1)/3 = 5N/6 - 1/3$.
For typical vision transformer sequences
($N = 197$ for ViT/16 on $224{\times}224$),
$s_i \approx 164 \gg 1$.
Since $\tilde{\mathbf{A}}_{cf}[i,:] = s_i \cdot \mathbf{A}_{cf}[i,:]$
where $\mathbf{A}_{cf} = \mathrm{Norm}(\tilde{\mathbf{A}}_{cf})$
is the row-normalised version, the unnormalised counterfactual
feature is:
\begin{equation}
  \tilde{\mathbf{h}}_{cf}
  = \tilde{\mathbf{A}}_{cf}\,\mathbf{V}
  = s_i\,\mathbf{A}_{cf}\,\mathbf{V}
  = s_i\,\mathbf{h}_{cf}.
  \label{eq:hcf_scale}
\end{equation}
The unnormalised loss therefore expands as:
\begin{align}
  \tilde{\mathcal{L}}_{\mathrm{CAFC}}
  &= \bigl\|\,\mathbf{h}_L - s_i\,\mathbf{h}_{cf}\,\bigr\|_2^2
  \nonumber\\[2pt]
  &= \|\mathbf{h}_L\|^2
     - 2\,s_i\,\langle\mathbf{h}_L,\,\mathbf{h}_{cf}\rangle
     + s_i^2\,\|\mathbf{h}_{cf}\|^2.
  \label{eq:loss_decomp}
\end{align}
The term $s_i^2\,\|\mathbf{h}_{cf}\|^2$ grows as
$\mathcal{O}(N^2)$ and depends only on the scalar row-sum
magnitude $s_i$, not on the attention direction.
Minimising $\tilde{\mathcal{L}}_{\mathrm{CAFC}}$ therefore
drives $s_i \to 0$~(scale collapse) rather than pulling
$\mathbf{h}_L$ toward $\mathbf{h}_{cf}$, and the gradient
with respect to model parameters is dominated by this
scale term, providing no useful signal about directional
feature consistency.

\paragraph{Normalisation restores the intended objective.}
After applying $\mathrm{Norm}(\cdot)$, all four maps
are valid probability distributions over the $N$ tokens
with unit row sums.
The counterfactual feature $\mathbf{h}_{cf} =
\mathbf{A}_{cf}\,\mathbf{V}$ is a proper weighted average
of the value rows under a maximally uninformative attention
distribution.
$\mathcal{L}_{\mathrm{CAFC}} =
\|\mathbf{h}_L - \mathbf{h}_{cf}\|_2^2$ then measures
the change in the feature vector solely due to the difference
in attention \emph{direction} between the learned and
counterfactual maps, with no scale confound.
This is the quantity that directly reflects attention-induced
feature instability under distribution shift and constitutes
an appropriate training signal for domain-robust feature
learning.

\section{Robustness Properties of the DGASS Threshold}

Equation~(15) in the main paper defines the DGASS threshold as the median of the
per-task minimum entropy:
\[
\lambda =
\operatorname{median}_{t\in\mathcal{T}}
\left\{
\min\left(
\bar H(f_s^1;X_t),
\bar H(f_s^2;X_t)
\right)
\right\}.
\]

We motivate this choice as a robust, label-free thresholding rule.

\textbf{Median as an $\ell_1$-robust statistic.}
For any finite set of scalar values $\{x_i\}_{i=1}^{n}$, any median
minimizes the sum of absolute deviations \cite{huber2011robust, law1986robust}:
\[
\operatorname{median}(\{x_i\})
\in
\arg\min_c \sum_{i=1}^{n} |x_i-c|.
\]
In contrast, the arithmetic mean minimizes squared deviations and is
therefore more sensitive to extreme entropy values. This distinction is
important for cross-task entropy statistics, where transfer tasks can have
substantially different domain-gap severity. Using the median gives each
task equal rank-based influence and provides an outlier-resistant operating
point for separating relatively low-gap and high-gap transfers.

\textbf{Distribution-free validity.}
The median does not require a parametric assumption about the shape of the
cross-task entropy distribution. This is appropriate for DGASS because the
threshold must be computed without target labels or target validation
accuracy. Parametric alternatives, such as fitting a Gaussian model or
choosing a distribution-specific percentile, would require estimating
additional distributional parameters from the same unlabeled entropy
statistics. The median rule avoids this extra modeling assumption while
remaining fully label-free.
\begin{table}[hb]
\centering
\scalebox{1}{
\begin{tabular}{lcccccc}
\toprule
$\lambda$ & 0.0 & 0.2 & 0.78 & 0.8 & 1.6 & 1.7 \\
\midrule
Avg. & 90.01 & 90.01 &{91.65} & 91.53 & 91.53 & 91.46 \\
\bottomrule
\end{tabular}}
\caption{\small Sensitivity of DGA-CAFC to the DGASS threshold
$\lambda$ on Office31. Accuracy remains stable over a broad
range around the selected value.}
\label{tab:lambda}
\end{table}

\textbf{Empirical robustness as complementary evidence.}
The threshold-sensitivity analysis in Table~\ref{tab:lambda}
shows that Office31 accuracy varies by less than $0.19$ percentage points
for $\lambda\in[0.78,1.7]$, a range that includes the median-derived value.
This indicates that DGASS does not require a precisely tuned threshold in
this setting. The role of the median rule is therefore not to identify a
single sharply optimal value, but to provide a reliable validation-free
operating point within a stable range of threshold values.

\begin{table}[ht]
    \begin{center} 
        \scalebox{0.9}{%
        \begin{tabular}{c|cccccc}
            \hline
            Model& A$\rightarrow$D & A$\rightarrow$W & D$\rightarrow$A & D$\rightarrow$W & W$\rightarrow$A & W$\rightarrow$D \\
            \hline
            \(f_s^1\) & 0.78 & 0.76 & 1.78 & 0.49 & 1.63 & 0.27 \\
            \(f_s^2\)           & 0.75 & 0.73 & 1.83 & 0.50 & 1.60 & 0.29 \\
            \hline
        \end{tabular}
        }
    \end{center}
    \caption{Entropy values for different domain adaptation tasks on the Office-31 dataset}
    \label{tab:office31_entropy}
\end{table}

\begin{table*}[ht]
\centering
\small
\begin{tabular}{l cccccc cccccc}
\hline
Model
  & A$\to$C & A$\to$P & A$\to$R
  & C$\to$A & C$\to$P & C$\to$R
  & P$\to$A & P$\to$C & P$\to$R
  & R$\to$A & R$\to$P & R$\to$C \\
\hline
$f_s^1$ & 1.32 & 1.08 & 0.92 & 1.27 & 1.02 & 1.03
        & 1.26 & 1.35 & 0.80 & 0.83 & 1.03 & 0.53 \\
$f_s^2$ & 1.25 & 0.97 & 0.84 & 1.25 & 1.03 & 1.01
        & 1.23 & 1.29 & 0.80 & 0.82 & 1.01 & 0.51 \\
\hline
\end{tabular}
\caption{Prediction entropy values for Office-Home transfer tasks.}
\label{tab:officehome_entropy}
\end{table*}

\begin{table*}[ht]
\centering
\small
\begin{tabular}{l cccccc cccccc}
\hline
Model
  & S$\to$C & S$\to$P & S$\to$R
  & C$\to$S & C$\to$P & C$\to$R
  & P$\to$S & P$\to$C & P$\to$R
  & R$\to$S & R$\to$C & R$\to$P \\
\hline
$f_s^1$ & 0.53 & 0.56 & 0.48 & 0.91 & 1.22 & 1.00
        & 0.50 & 0.55 & 0.39 & 0.33 & 0.31 & 0.27 \\
$f_s^2$ & 0.54 & 0.57 & 0.49 & 0.94 & 1.23 & 1.02
        & 0.48 & 0.52 & 0.38 & 0.32 & 0.29 & 0.25 \\
\hline
\end{tabular}
\caption{Prediction entropy values for DomainNet40 transfer tasks.}
\label{tab:domainnet40_entropy}
\end{table*}

\begin{table}[ht]
    \begin{center} 
        \scalebox{0.75}{%
        \begin{tabular}{c|cccccc}
            \hline
            Model& A$\rightarrow$D & A$\rightarrow$W & D$\rightarrow$A & D$\rightarrow$W & W$\rightarrow$A & W$\rightarrow$D \\
            \hline
           Domain Gap from \(f_s^1\) & 0.78 & 0.76 & 1.78 & 0.49 & 1.63 & 0.27 \\
            \hline
        \end{tabular}
        }
    \end{center}
    \caption{Entropy values (Domain Gap) for different adaptation tasks on Office-31 from \(f_s^1\) model}
    \label{tab:office31_entropy_from_fs1}
\end{table}

\begin{table}[ht]
    \begin{center} 
        \scalebox{1}{%
        \begin{tabular}{c|c}
            \hline
            Task& Model Selected \\
            \hline
          $A \rightarrow D$  & \(f_s^1\) \\
          $A \rightarrow W$  & \(f_s^2\) \\
          $D \rightarrow A$  &  \(f_s^2\)\\
          $D \rightarrow W$  &  \(f_s^1\)\\
          $W \rightarrow A$  &  \(f_s^1\)\\
          $W \rightarrow D$  & \(f_s^1\) \\
            \hline
        \end{tabular}
        }
    \end{center}
    \caption{Model selection through DGASS module on office31 dataset.}
    \label{tab:office31_model_selection}
\end{table}

\section{Entropy Values for Model selection}
\label{sec:Entropy_Values_for_Model_selection}
The entropy values utilized for model selection is reported in Table \ref{tab:office31_entropy}, \ref{tab:officehome_entropy} and \ref{tab:domainnet40_entropy} for office31, office-home and domainnet40 dataset respectively. For each dataset we report the entropy values from both variants of the source model $f_s^1$ and $f_s^2$.

\section{Example of Model Selection}
\label{sec:Example_of_Model_Selection}
Example of model selection on office31 dataset. First step is to analyze the domain gap for each task based on the entropy from any of the two models \(f_s^1\) and \(f_s^2\). Here, for office31 dataset, we report the prediction entropy in Table \ref{tab:office31_entropy_from_fs1}. The threshold $\lambda =0.78$\\

\begin{table*}[ht]
\begin{center} 

    \begin{tabular}{c|cccccccccccc|c}
    \toprule
       Method  & A$\rightarrow$C & A$\rightarrow$P & A$\rightarrow$R & C$\rightarrow$A & C$\rightarrow$P & C$\rightarrow$R & P$\rightarrow$A & P$\rightarrow$C & P$\rightarrow$R & R$\rightarrow$A & R$\rightarrow$C & R$\rightarrow$P & Avg. \\
         \midrule
        Variant-1     & 68.77 & 81.78 & 85.36 & 75.94 & 85.22 & 85.24 & 77.79 & 67.29 & 86.55 & 80.06 & 70.03 & 88.53 & 79.40 \\
        
        Variant-2        & 65.77 & 82.29 & 84.53 & 73.18 & 82.32 & 83.34 & 76.68 & 66.78 & 85.91 & 79.52 & 68.09 & 89.07 & 78.12 \\
        Variant-3     & 67.06 & 82.07 & 85.01 & 75.36 & 83.49 & 83.93 & 76.64 & 66.64 & 86.00 & 79.69 & 69.03 & 88.60 & 78.62 \\
        

          \bottomrule
    \end{tabular}

\end{center}
    \caption{\small Generalization ability of three variants of source model on Office-Home dataset.}
    \label{tab:source_variant_ablation}
\end{table*}

\noindent Tasks with high domain gap are  ($A \rightarrow D$, $D \rightarrow A$ and $W \rightarrow A$).\\
Tasks with low domain gap are  ($A \rightarrow W$, $D \rightarrow W$ and $W \rightarrow D$).

\noindent Now, if the domain gap is higher than or equal to threshold 0.78 (High domain gap tasks), select high entropy model otherwise select low entropy model based on Table \ref{tab:office31_entropy}.\\ The results are reported in Table \ref{tab:office31_model_selection}

\section{Design Rationale for Counterfactual Perturbations}

The three perturbations in Eq.~(4) are designed to target structurally
distinct failure modes of attention under distribution shift. Inspired by
counterfactual attention interventions used in prior attention-learning
work~\cite{rao2021counterfactual,yu2025attention}, we use a small set of
complementary perturbations rather than searching over the full space of
possible attention corruptions. These perturbations stress-test different
forms of attention instability that can arise when source-learned attention
patterns are applied to shifted target domains.

\textbf{Uniform attention.}
$A_{\mathrm{unif}}$ represents the no-focal-structure limit: attention is
spread uniformly across tokens, testing whether the downstream feature is
overly dependent on a concentrated learned attention pattern.

\textbf{Random attention.}
$A_{\mathrm{rand}}$ represents unstructured corruption: attention weights are
replaced with a random row-normalized distribution, testing robustness to
noisy or unreliable attention allocation.

\textbf{Reversed attention.}
$A_{\mathrm{rev}}$ represents complementary saliency: high-attention regions
under the learned map are suppressed while lower-attention regions are
emphasized, testing whether the representation over-relies on source-salient
regions that may not transfer under domain shift.

Together, these perturbations cover three qualitatively different regimes:
removing focal structure, injecting unstructured noise, and emphasizing
regions complementary to the learned attention. We do not claim that they
exhaust all possible attention corruptions or form an optimal perturbation
basis. Instead, they provide a compact and label-free set of attention-space
stress tests for regularizing attention-conditioned features.

\section{Ablation study of the effectiveness of source model variants} 
\label{Ablation study of the effectiveness of source model variants}
Analysis of how different source model variants, trained with distinct attention modules impact the overall adaptation performance. For fair analysis we exclude the CAFC and DGASS modules from the adaptation framework. We train three variants of source model and adapt it on target domain by simply leveraging self-labeling and information maximization loss. Variants are created by utilizing different attention softmax against each variant, on office-home.
Variant-1 is trained with standard softmax-based attention weights ($\alpha_{ij}^1$), variant-2 is trained with a weighted softmax ($\alpha_{ij}^2$) and variant-3 is trained with learnable weighted average of variant-1 and variant-2 ($\alpha_{ij}^3$) as shown in main paper.
As reported in Table \ref{tab:source_variant_ablation}, different variants shows its generalizable ability. We select the best two variant which is Variant-1 and Variant-3 in this case. Based on this study, we utilized these two variants across all datasets.  

\section{Statistical Significance and Seed-Level Robustness}
\label{sec:statistical_robustness}

To assess whether the improvements introduced by CAFC are robust to random-seed variation, we perform paired significance analysis using matched seed-level runs under the standard DomainBed-style ImageNet-pretrained ResNet-50 protocol. For each DG objective, the baseline model and its corresponding CAFC-augmented variant are evaluated using the same random seed. We conduct the analysis over four objectives for which complete matched seed-level results are available on both PACS and VLCS: CORAL, ERM, GroupDRO, and Mixup. This gives $4$ objectives $\times$ $3$ seeds $=12$ matched seed-objective observations per dataset.

For each matched observation $i$, we define the accuracy difference as
\begin{equation}
\Delta_i =
\mathrm{Acc}_{\mathrm{Base+CAFC},i}
-
\mathrm{Acc}_{\mathrm{Base},i}.
\end{equation}
Positive $\Delta_i$ indicates an improvement from adding CAFC.

We report the mean paired improvement, its 95\% confidence interval, a two-sided paired $t$-test, and a one-sided Wilcoxon signed-rank test with the alternative hypothesis that CAFC improves accuracy. The paired $t$-test measures whether the mean matched improvement differs from zero, while the Wilcoxon test provides a complementary non-parametric assessment that does not rely on normality of the paired differences.

\begin{table}[t]
\centering

\resizebox{\linewidth}{!}{
\begin{tabular}{c l c c r}
\toprule
Seed & Objective & Base & +CAFC & $\Delta$ \\
\midrule
0 & CORAL    & 80.92 & 82.55 & +1.63 \\
0 & ERM      & 78.47 & 80.68 & +2.21 \\
0 & GroupDRO & 82.21 & 81.81 & -0.41 \\
0 & Mixup    & 84.93 & 85.50 & +0.57 \\
\midrule
1 & CORAL    & 81.78 & 82.48 & +0.70 \\
1 & ERM      & 78.07 & 81.03 & +2.96 \\
1 & GroupDRO & 79.23 & 81.86 & +2.62 \\
1 & Mixup    & 82.31 & 83.06 & +0.76 \\
\midrule
2 & CORAL    & 81.80 & 81.78 & -0.02 \\
2 & ERM      & 80.82 & 82.70 & +1.88 \\
2 & GroupDRO & 82.86 & 85.17 & +2.31 \\
2 & Mixup    & 83.66 & 85.07 & +1.41 \\
\bottomrule
\end{tabular}}
\caption{Matched seed-level comparison on PACS under the standard DomainBed-style ImageNet-pretrained ResNet-50 protocol. Each Base and +CAFC value is the average accuracy across the four held-out PACS domains for the corresponding objective and random seed. $\Delta$ denotes the paired gain from CAFC.}

\label{tab:pacs_seed_significance}
\end{table}

\begin{table}[t]
\centering

\resizebox{\linewidth}{!}{
\begin{tabular}{c l c c r}
\toprule
Seed & Objective & Base & +CAFC & $\Delta$ \\
\midrule
0 & CORAL    & 77.66 & 79.35 & +1.68 \\
0 & ERM      & 77.06 & 79.87 & +2.81 \\
0 & GroupDRO & 77.18 & 78.53 & +1.35 \\
0 & Mixup    & 79.15 & 79.67 & +0.52 \\
\midrule
1 & CORAL    & 77.84 & 78.10 & +0.26 \\
1 & ERM      & 77.06 & 80.82 & +3.77 \\
1 & GroupDRO & 77.59 & 79.42 & +1.83 \\
1 & Mixup    & 78.85 & 80.28 & +1.43 \\
\midrule
2 & CORAL    & 77.66 & 79.19 & +1.54 \\
2 & ERM      & 77.77 & 80.58 & +2.81 \\
2 & GroupDRO & 77.05 & 78.33 & +1.28 \\
2 & Mixup    & 78.48 & 79.11 & +0.64 \\
\bottomrule
\end{tabular}}
\caption{Matched seed-level comparison on VLCS under the standard DomainBed-style ImageNet-pretrained ResNet-50 protocol. Each Base and +CAFC value is the average accuracy across the four held-out VLCS domains for the corresponding objective and random seed.}

\label{tab:vlcs_seed_significance}
\end{table}

\begin{table}[t]
\centering

\resizebox{\linewidth}{!}{
\begin{tabular}{l c c c c c}
\toprule
Dataset &
Improved pairs &
Mean $\Delta$ &
95\% CI &
Paired $t$-test &
Wilcoxon \\
\midrule
PACS &
10/12 &
+1.38 pp &
[0.71, 2.06] &
$t(11)=4.48,\ p=0.00092$ &
$p=0.00122$ \\
VLCS &
12/12 &
+1.66 pp &
[1.00, 2.31] &
$t(11)=5.56,\ p=0.00017$ &
$p=0.00024$ \\
\bottomrule
\end{tabular}}
\caption{Summary of paired statistical significance tests. Tests use the 12 matched seed-objective observations per dataset. The paired $t$-test is two-sided; the Wilcoxon signed-rank test uses the one-sided alternative that CAFC improves accuracy.}
\label{tab:statistical_significance_summary}
\end{table}

\paragraph{PACS.}
As shown in Table~\ref{tab:pacs_seed_significance}, CAFC improves 10 of the 12 matched seed-objective observations. The mean paired gain is $+1.38$ percentage points, with a 95\% confidence interval of $[0.71,2.06]$. A paired $t$-test yields
$t(11)=4.48$ with $p=0.00092$, indicating that the mean improvement is significantly different from zero. The result is corroborated by the one-sided Wilcoxon signed-rank test ($p=0.00122$). Although two individual paired runs show negligible or modest negative changes, the aggregate effect across objectives and seeds remains consistently positive and statistically significant.

\paragraph{VLCS.}
Table~\ref{tab:vlcs_seed_significance} shows an even stronger consistency pattern on VLCS: CAFC improves all 12 matched seed-objective observations. The mean paired improvement is $+1.66$ percentage points, with a 95\% confidence interval of $[1.00,2.31]$. The paired $t$-test gives
$t(11)=5.56$ and $p=0.00017$, while the one-sided Wilcoxon signed-rank test gives $p=0.00024$. Both tests therefore support a robust positive effect of CAFC across objectives and random seeds.
\begin{figure*}[ht]
    \centering
    \scalebox{1}{
    \begin{minipage}{0.45\linewidth}
        \centering
        \includegraphics[width=\linewidth]{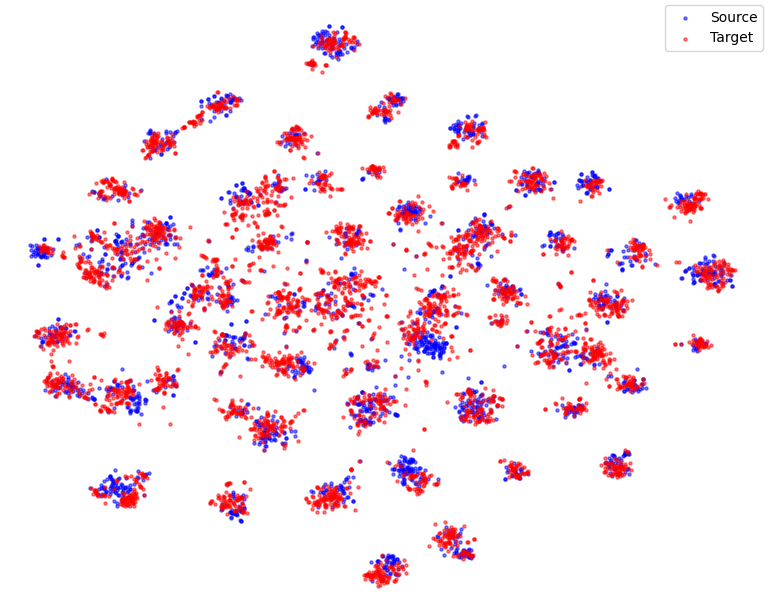}
        \label{fig:tsne_office_DA_non_adapted}
    \end{minipage}%
    \hfill
    \begin{minipage}{0.45\linewidth}
        \centering
        \includegraphics[width=\linewidth]{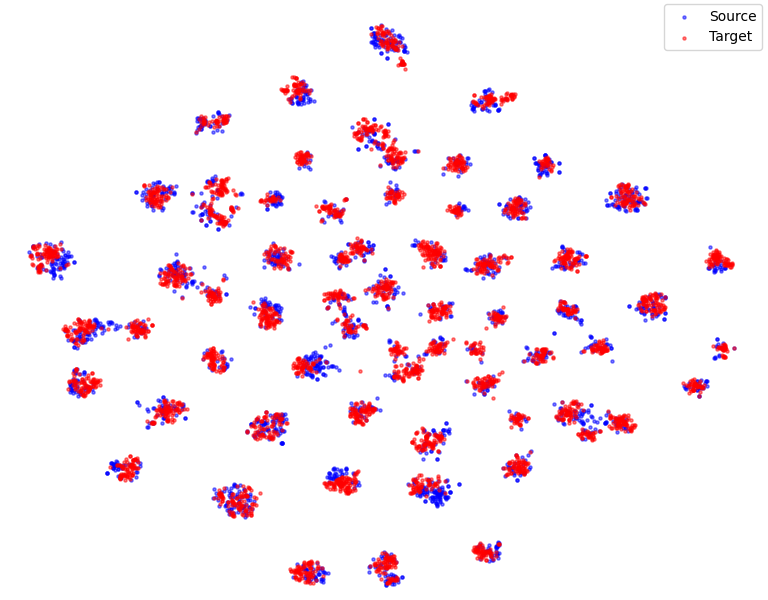}
        \label{fig:tsne_office_DA_adapted}
    \end{minipage}
    }
    \caption{\small t-SNE visualization of the $A \rightarrow R$ of OfficE-Home dataset. Left: Source-only. Right: DGA-CAFC (best view in color).}
    \label{fig:tsne_office_DA}
\end{figure*}

\paragraph{Discussion.}
These paired seed-level results complement the mean$\pm$standard-deviation results reported in the main paper. In particular, the gains from CAFC are not driven by a single objective or an isolated random seed: CAFC improves 10/12 matched runs on PACS and 12/12 on VLCS, with statistically significant paired improvements on both benchmarks. This strengthens the evidence that the observed DG gains under the standard DomainBed protocol reflect a systematic effect of the CAFC regularizer rather than random seed variation.


        
        



\section{Seed Variance Reduction}
\label{sec:variance_reduction}

Beyond mean accuracy gains, CAFC reduces seed-to-seed variance for
several DG objectives. Table~\ref{tab:variance_reduction} reports the
standard deviation across three seeds, before and after adding CAFC,
under the ResNet-18-from-scratch isolation setting (main paper,
Tables 4--5).

\begin{table}[t]
\centering
\resizebox{\linewidth}{!}{
\begin{tabular}{l l c c}
\toprule
Dataset & Objective & Std (Base) & Std (+CAFC) \\
\midrule
PACS & GroupDRO & $\pm$1.85 & $\pm$1.06 \\
PACS & Mixup    & $\pm$5.52 & $\pm$2.46 \\
VLCS & GroupDRO & $\pm$1.79 & $\pm$0.15 \\
VLCS & Mixup    & $\pm$0.89 & $\pm$0.36 \\
\bottomrule
\end{tabular}}
\caption{Seed-to-seed standard deviation, before and after CAFC, for the
two objectives showing the largest variance reduction (ResNet-18
isolation setting). Lower values indicate more consistent performance
across random seeds.}
\label{tab:variance_reduction}
\end{table}

For both objectives on both datasets, CAFC substantially tightens the
spread of results across seeds -- most notably a $\sim$12$\times$
reduction for GroupDRO on VLCS ($\pm$1.79 $\to$ $\pm$0.15). This
indicates that CAFC's benefit is not limited to the mean accuracy gains
reported in the main paper, but also improves the reliability and
predictability of training outcomes across random initializations.

\section{t-SNE visualization}
\label{sec:t-SNE}
Figure \ref{fig:tsne_office_DA} shows the t-SNE visualization of source (blue) and target features (red). Left one shows the learned features of source and target with the source-only model, right one shows the learned features through the proposed DGA-CAFC model for $A \rightarrow R$ task.

\end{document}